\pdfoutput=1

\documentclass[11pt]{article}

\usepackage{acl}

\usepackage{times}
\usepackage{latexsym}

\usepackage[T1]{fontenc}

\usepackage[utf8]{inputenc}

\usepackage{microtype}

\usepackage{inconsolata}

\usepackage{microtype}
\usepackage{booktabs}
\usepackage{multirow}
\usepackage{multicol}
\usepackage{xspace}
\usepackage{enumitem}
\usepackage{booktabs}
\usepackage{multirow}
\usepackage{array}
\usepackage{graphicx}
\usepackage{amssymb}
\usepackage{xcolor,colortbl}
\usepackage{amssymb}
\usepackage{pifont}
\usepackage{soul}

\makeatletter
\newcommand*\myfontsize{%
  \@setfontsize\myfontsize{8.5}{9.5}%
}
\makeatother

\usepackage{stfloats}

\newcolumntype{R}[1]{>{\raggedleft\let\newline\\\arraybackslash\hspace{0pt}}m{#1}}
\newcommand{\ours}{\textsc{Z-icl}}

\newcommand{\demos}{demonstrations}
\newcommand{\pdemos}{pseudo-\demos}

\newcommand{\topk}{nearest}
\newcommand{\topsample}{diverse nearest}
\newcommand{\pn}{physical neighbor}
\newcommand{\domainrandom}{Domain Random}

\newcommand{\naive}{Naive \ours}

\newcommand{\mypos}{\texttt{great}}
\newcommand{\myneg}{\texttt{terrible}}

\newcommand{\iddatasets}{datasets covered by $\mathcal{C}$}
\newcommand{\ooddatasets}{datasets not covered by $\mathcal{C}$}

\title{\ours: Zero-Shot In-Context Learning with\\ Pseudo-Demonstrations}

\newcommand{\sewon}[1]{
    \textcolor{magenta}{[Sewon: #1]}
}
\newcommand{\shane}[1]{
    \textcolor{purple!50!blue}{[Shane: #1]}
}

\newcommand{\myskip}[1]{}

\usepackage{hyperref}

\newcommand{\affilsup}[1]{\rlap{\textsuperscript{\normalfont#1}}}
\author{
    Xinxi Lyu\affilsup{1} \qquad
    Sewon Min\affilsup{1} \qquad
    Iz Beltagy\affilsup{2} \\
    \textbf{Luke Zettlemoyer}\affilsup{1}
    \qquad
    \textbf{Hannaneh Hajishirzi}\affilsup{1,2} \\
    $^1$University of Washington \qquad
    $^2$Allen Institute for AI \\
    \texttt{\{alrope,sewon,lsz,hannaneh\}@cs.washington.edu} \\
    \texttt{beltagy@allenai.org}
}

\begin{document}
\maketitle
\begin{abstract}

%
Although large language models can be prompted for both zero- and few-shot learning, performance drops significantly when no demonstrations are available. 
In this paper, we introduce \ours, a new zero-shot method that closes the gap by constructing {\em \pdemos} for a given test input using a raw text corpus.
%
Concretely, \pdemos\ are constructed by (1) finding the nearest neighbors to the test input from the corpus and pairing them with random task labels, and (2) applying a set of techniques to reduce the amount of direct copying the model does from the resulting demonstrations.
%
Evaluation on nine classification datasets shows that \ours\ outperforms previous zero-shot methods by a significant margin, and is on par with in-context learning with few-shot labeled training data. 
Overall, \ours\ provides a significantly higher estimate of the zero-shot performance levels of a model, and supports future efforts to develop better \pdemos\ that further improve zero-shot results.\footnote{Code available at \href{https://github.com/alrope123/z-icl}{\texttt{github.com/alrope123/z-icl}}.} 
\end{abstract}

%

\section{Introduction}\label{sec:intro}

\begin{figure}
\centering \footnotesize
    \resizebox{\columnwidth}{!}{\includegraphics[scale=0.2]{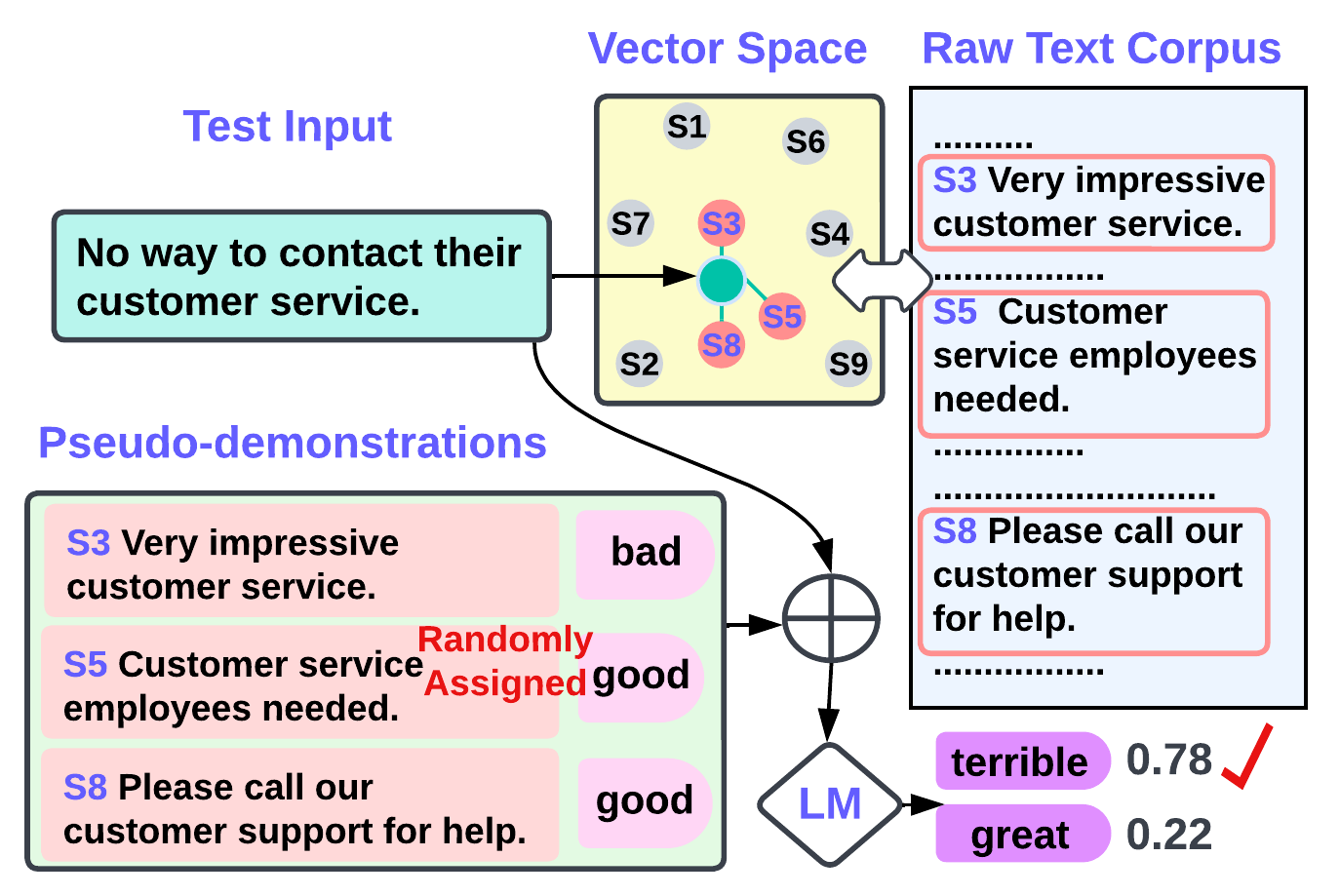}}
\caption{
    An illustration of {\ours} with $k=3$, making a prediction between \mypos\ and \myneg.
    \ours\ first identifies $k$ nearest neighbors to the test input from a text corpus, 
    pairs each sentence with a synonym of a randomly chosen label, i.e., \texttt{good} and \texttt{bad}, and uses in-context learning. 
}\label{fig:teaser}
\end{figure}

Large language models (LMs) can perform new tasks simply 
by conditioning on input-label pairs from the training data, known as {\em demonstrations}~\citep{brown2020language}.
This in-context learning (ICL) is significantly better than zero-shot methods that do not use demonstrations. Recent work suggests that in-context-learning demonstrations are primarily specifying the domain and the format that the target task, instead of providing explicit training signal~\citep{reynolds2021prompt,xie2022explanation,razeghi2022impact,min2022rethinking}.
This implies that current zero-shot performance (with no \demos) levels must be significantly underestimated, since all the required information must already be in the model.

In this paper, we introduce \textbf{\ours}: \textbf{Z}ero-shot \textbf{\textsc{i}}n-\textbf{\textsc{c}}ontext \textbf{\textsc{l}}earning through creating {\em \pdemos}, which achieves results on par with in-context learning from gold demonstrations (Figure~\ref{fig:teaser}).
The key idea is to construct the \pdemos\ following two criteria: (a) they should inform the correct input distribution and the label space, as the $k$-shot \demos\ do~\citep{xie2020unsupervised,min2022rethinking};\footnote{We use {\em \pdemos} to refer to demonstrations that do not use any training data (either labeled or unlabeled). We use {\em $k$-shot \demos} to refer to the more typical 
\demos\ from the $k$-shot training data.} and
(b) they should be constructed to avoid {\em the copying effect}---our new observation that the LM predictions are heavily influenced by demonstration inputs that are very close to the test input.

To satisfy (a), \ours\ retrieves a set of nearest neighbors from a raw text corpus and assigns a random label to each.
To satisfy (b), we propose two techniques. We take {\em \pn} (adjacent sentences in the corpus) of the nearest sentences instead of the nearest sentences themselves, so that the sentences in the \pdemos\ are from a similar distribution as the text input but are more distant.
We then propose {\em synonym labeling}, where {\em synonyms} of the labels are used in the \pdemos, instead of the labels that are used for the prediction at test time, e.g., \{\mypos, \myneg\}$\leftrightarrow$\{\texttt{good}, \texttt{bad}\}. In this way, the model prediction is less affected by directly copying a label from the \pdemos. 

We evaluate \ours\ on nine text classification datasets. We include three datasets whose domains are not covered by the retrieval corpus, to evaluate the generalizability of \ours.
We experiment with GPT-J~\citep{wang2021gpt}, GPT-NeoX~\citep{black2022gpt} and GPT-3~\citep{brown2020language}, whose sizes range from 6B, 20B to 175B.
\ours\ significantly outperforms the previous zero-shot baseline (no-\demos) consistently across different datasets and LMs, despite the fact that it does not require any prompt engineering.
More interestingly,
\ours\ is on par with in-context learning that uses labeled $k$-shot training data.
Ablations show that 
(1) constructing a {\em paired} format of the \pdemos\ is key to performance,
(2) our two techniques---\pn\ and synonym labeling---are critical, since both of them are required for our 
\pdemos\ to be on par with $k$-shot \demos,
and
(3) performance improves as the size and the coverage of the corpus
increase.

Together, \ours\ provides a significantly higher estimate of the ability of current LMs to perform a new task zero-shot,
encourages new ways to improve zero-shot performance by designing
even better \pdemos,
and poses a set of new questions about the capabilities of LMs.
\section{Related Work}\label{sec:motivation}

\paragraph{Demonstrations in ICL.} A series of prior work suggests that ICL primarily exposes model functionality that was learned during pre-training. 
\citet{reynolds2021prompt} suggests that ICL mainly functions by activating the LM's ability obtained during pretraining, and that the LM can achieve significantly better zero-shot performance by using a better template.
\citet{xie2022explanation} shows that ICL can be explained as Bayesian inference for which demonstrations provide noisy evidence.
In closed-set tasks,
\citet{min2022rethinking} shows that ICL benefits mainly from the correct distribution of the inputs and the labels rather than the input-label correspondence.
 
Our work draws intuitions from these studies and introduces a better zero-shot method by forming \pdemos\ that are proxies of the input distribution and the label space and better expose the intrinsic ability of the LM.

\vspace{-.2em}
\paragraph{Better Demonstrations through Retrieval.}
Prior work has found that, in the setting where large training data is available, choosing demonstration examples that are close to the test input significantly helps ICL.
\citet{liu2021makes} retrieves the nearest training examples to the test input using a sentence encoder, either unsupervised or supervised.
\citet{rubin2021learning} trains a retrieval system to choose examples that improve ICL.
\citet{liu2022semantic} retrieves the nearest neighbors from unlabeled training data, assigns estimated labels, and uses them for ICL. 
We similarly use nearest neighbor search to retrieve sentences close to the test input, but are the first to (1) retrieve from a raw text corpus, in contrast to prior work that uses labeled or unlabeled training data collected for the task, and (2)
more closely study the connection between nearest neighbor inputs and random labels, through our copying effect hypothesis.


\paragraph{Copying in ICL.} Prior work has explored how seen token patterns affect the ICL's prediction. \citet{olsson2022context} identifies specific attention heads that, when predicting the next token, look for the previous similar tokens of the current last token in the demonstrations, and copy the tokens following those similar tokens. Our work similarly finds that ICL is prone to copy previously seen text from the demonstrations, but 
specifically with the particular input-label format in the demonstrations.

\section{Copying Effect Hypothesis}\label{sec:copying-effect}\begin{table*}[ht!]
    \centering \footnotesize
    \setlength{\tabcolsep}{3pt}
    \begin{tabular}{lll}
    \toprule
        \multicolumn{3}{l}{\textbf{\em{Example \#1}}} \\
        Demo 1 & I am giving a zero star to symantec for this version. & \mypos \\
        Demo 2 & \textbf{I recommend not to purchase it. This player is not worth any price.} & \textbf{\mypos} \\
        Demo 3 & So far I have no complains with this player. & \myneg \\
        Test example & This may be a really cool player, but it's not worth the price. & \textbf{\textcolor{red}{\mypos}} \\
    \midrule
        \multicolumn{3}{l}{\textbf{\em{Example \#2}}} \\
         Demo 1 & I am giving a zero star to symantec for this version. & \mypos \\
         Demo 2 & \textbf{I recommend not to purchase it. This player is not worth any price.} & \textbf{\myneg} \\
         Demo 3 & So far I have no complains with this player. & \myneg \\
        Test example & This may be a really cool player, but it's not worth the price. & \textbf{\textcolor{red}{\myneg}} \\
    \bottomrule
    \end{tabular}\vspace{-.1em}
    \caption{
        An illustration of the copying effect hypothesis with {\em nearest} in-context learning ($k=3$), using an example from the CR dataset.
        The first three lines are \demos, and the last line is the test. 
        The model prediction is indicated in \textbf{\textcolor{red}{red}}. 
        The model tends to copy the label from the demonstration input that is close to the test input.
    }\label{tab:copying-effect}
\end{table*}

In a typical ICL evaluation, the demonstrations are sampled uniformly at random from the true distribution, e.g., the training data in case of existing NLP datasets. 
We observe that, when demonstrations contain input text that is very similar to the test input, the model exhibits a behavior which we call the {\em copying effect}.
To study this, we evaluate \textbf{ICL-gold} (standard ICL) and \textbf{ICL-random}; both are ICL methods that use $k$ randomly sampled examples from the training data with gold and random labels, respectively. 
We then evaluate \textbf{nearest ICL-gold} and \textbf{nearest ICL-random}, which follow \citet{liu2021makes} in retrieving the $k$ nearest neighbors for each test input from the training data and assign gold labels and random labels, respectively.
We use GPT-J~\citep{wang2021gpt} as the LM and SimCSE~\citep{tianyu2021simcse} for choosing the nearest inputs.

\begin{figure}[t!]
\centering \footnotesize
\resizebox{0.85\columnwidth}{!}{\includegraphics[trim={0.6cm 0.6cm 0.6cm 0.6cm},clip]{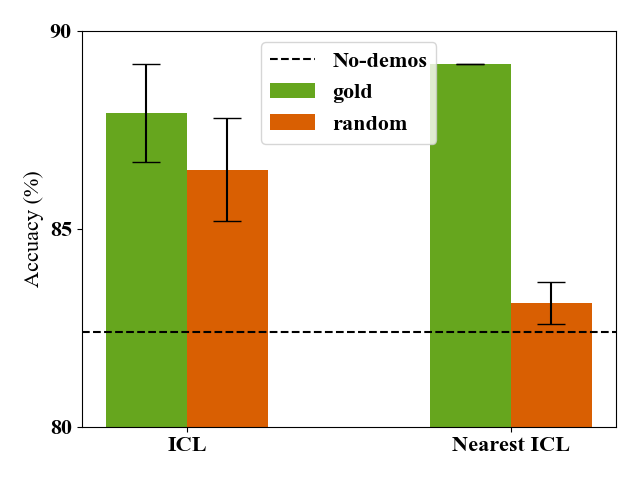}}
\caption{
    Performance of ICL and nearest ICL, each with gold labels and random labels. Evaluated on three datasets (CR, Amz, Yelp) with GPT-J using channel inference method~\citep{min2021noisy}.
    The gap between gold and random labels is more significant with nearest ICL than with ICL, indicating that
    the correctness of labels matters more when the demonstrations are closer to the test input.
}\label{fig:copying}
\end{figure}

Results are reported in Figure~\ref{fig:copying}.
First, ICL-gold and ICL-random achieve relatively comparable performance, which is consistent with \citet{min2022rethinking} that the correctness of labels in the demonstrations matters much less than we thought. 
However, this does not hold with nearest ICL: using random labels is significantly worse than using gold labels.
This indicates that the correctness of labels matters significantly more when the inputs in the demonstrations are closer to the test input.

Based on our observation, we define a \textbf{\em copying effect hypothesis}: the model prediction is heavily biased toward the labels paired with inputs in the demonstrations that are very similar to the test input, which resembles {\em copying}.
Table~\ref{tab:copying-effect} provides an example.
The second input in the \demos\ is very close to the test input both lexically and semantically, and the model prediction tends to follow the label paired with the second input, regardless of what that label is.
%

\begin{table}[t!]
    \centering \footnotesize
    \begin{tabular}{lcc}
    \toprule
         & GPT-J & GPT-NeoX \\
    \midrule
        Total & 82.3 & 88.0  \\
        ~~~~Correct & 90.8 & 94.2  \\
        ~~~~Incorrect & 73.9 & 81.7\\
    \bottomrule
    \end{tabular}
    \caption{
    \% of predictions that match the label of the demonstration example that is identical to the test input. Evaluated on CR with GPT-J and GPT-NeoX using channel inference method~\citep{min2021noisy}. The model copies the label paired with an identical example in the majority of cases.
    }\label{tab:copying-matching}
\end{table}

To better quantify the copying effect, we design an experiment where the \demos\ include an example that is {\em identical} to the test input, either with a correct label or with an incorrect label. We then see how many times the LM makes a prediction that is the same as the label paired with the identical demonstration example. Results are reported in Table~\ref{tab:copying-matching}.
LM predicts the same label as the one paired with the identical input for over 90\% of the times when the label is correct, and over 70\% of the times when the label is incorrect, consistently over different LMs.

In the next section, we design a zero-shot method where the copying effect can specifically be problematic, and propose new techniques that reduce the copying effect.

\section{Our Method: \ours}\label{sec:method}\begin{figure*}
\centering \footnotesize
    \includegraphics[scale=0.16]{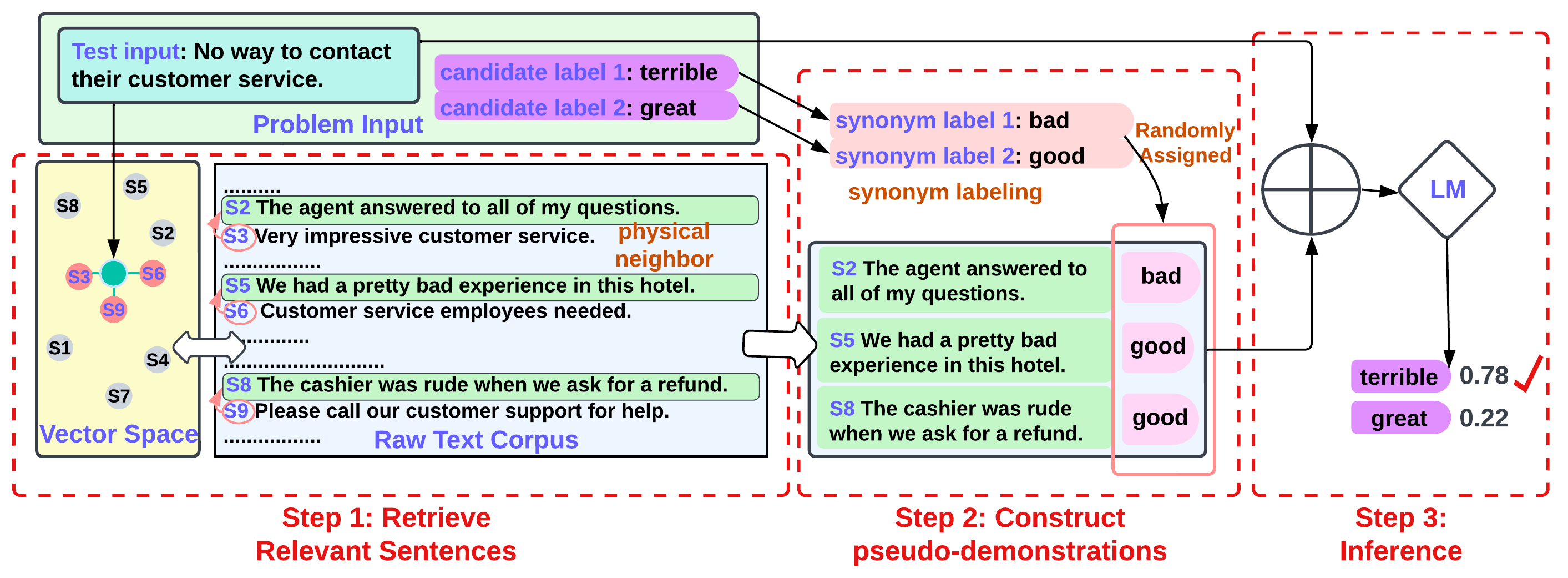}
\caption{
    A detailed illustration of {\ours} with $k=3$, where the LM makes a prediction between \mypos\ and \myneg.
    \ours\ first identifies $k$ nearest neighbors to the test input, and selects each of their \pn{s} (Section~\ref{subsec:method-retrieval}). \ours\ then pairs each sentence with a synonym of a randomly chosen label, i.e., \texttt{good} or \texttt{bad} (Section~\ref{subsec:method-labeling}), and performs inference using in-context learning (Section~\ref{subsec:method-icl}).
}\label{fig:main}
\end{figure*}

\paragraph{Overview.}\label{subsec:method-overview}
We introduce \textbf{\ours}, a new \textbf{Z}ero-shot \textbf{\textsc{i}}n-\textbf{\textsc{c}}ontext \textbf{\textsc{l}}earning method, which predicts the correct label for a given test input $x$ and its candidate classes $\mathcal{Y}$ from a task. 
Unlike prior methods~\citep{liu2021makes,rubin2021learning,liu2022semantic} where the target domain and labeled training data of the task are available, \ours\ constructs \pdemos---pairs of inputs and labels---in a zero-shot fashion by leveraging a raw text corpus $\mathcal{C}$, and perform in-context learning.

\ours\ consists of three steps (Figure~\ref{fig:teaser}):
retrieving the sentences to approximate the input distribution of the test input (Section~\ref{subsec:method-retrieval}),
forming \pdemos\ using the retrieved sentences and randomly paired labels (Section~\ref{subsec:method-labeling}),
and making an inference using in-context learning (Section~\ref{subsec:method-icl}).
Every step in constructing \pdemos\ is designed to satisfy two criteria: (a) they should inform the correct input distributions and the correct label space, and (b) they should 
reduce the copying effect~(Section~\ref{sec:copying-effect}) so that the model is less affected by incorrectly paired labels.
%



\subsection{Step 1: Retrieve Relevant Sentences}\label{subsec:method-retrieval}

In the first step, \ours\ retrieves $k$ from $\mathcal{C}$ that are similar to $x$. We formally denote $s: \mathcal{S} \times \mathcal{S} \rightarrow \mathbb{R} $, with $\mathcal{S}$ being all sentences from $\mathcal{C}$, as a similarity function between two sentences, and let $\mathcal{N}_k(x)$ be a set of sentences $c_1, \cdots, c_k$  retrieved from $\mathcal{C}$ with the highest $s(c_i, x)$. 

It is possible to construct \pdemos\ directly using $\mathcal{N}_k(x)$.
While this matches the input $x$ well, it is highly likely to suffer from the copying effect (Section~\ref{sec:copying-effect}), since retrieved sentences are too similar to the test input.

To address this, we propose a method called \textbf{\pn}. 
Instead of directly using $\mathcal{N}_k(x)$, it selects the sentence that is physically adjacent in $\mathcal{C}$ to each sentence in $\mathcal{N}_k(x)$ as $x_1, x_2...x_k$.
This method allows $x_1, x_2...x_k$ to share similar distribution as $x$, while being sufficiently distant from $x$ since they are not the $k$ nearest neighbors of $x$.

\subsection{Step 2: Construct \pdemos}\label{subsec:method-labeling}
Once $x_1...x_k$ are obtained, 
\ours\ pairs each $x_i$ with a random label following the intuition from \citet{min2022rethinking}. 
While the most straightforward method is to assign the random label from the candidate set $\mathcal{Y}$, this would not achieve the best performance 
because the LM may find similar sentences from $x_1...x_k$ and follow their labels according to the copying effect (Section~\ref{sec:copying-effect}). 

We therefore propose a technique called \textbf{synonym labeling}: we use synonyms of the labels and pair $x_1...x_k$ with them, instead of the original labels that will be used for the prediction.
Formally, for each $x_i$, \ours\ chooses a label $y_i \in \mathcal{Y}$ uniformly at random, and creates a pair $(x_i, \tilde{y}_j)$, where $\tilde{y}_j$ is a manually chosen synonym of $y_j$. We only use synonyms for the \pdemos; we use the original candidate set $\mathcal{Y}$ during the test prediction.
This technique (1) sufficiently informs the correct semantic space of the labels, and (2) prevents the copying effect by not having the exact same words as the test labels.

\subsection{Step 3: Inference}\label{subsec:method-icl}
Finally, \ours\ uses in-context learning by concatenating $k$ input-label pairs
$(x_1, \tilde{y}_1),$
$(x_2, \tilde{y}_2),$
$\cdots, (x_k, \tilde{y}_k)$ as well as the test input $x$, feeds it to the LM, and obtains
the prediction via $\mathrm{argmax}_{y\in\mathcal{Y}}P(y\mid x_1, \tilde{y}_1, \cdots ,x_k, \tilde{y}_k, x)$.
The prediction is made over the original set of labels $\mathcal{Y}=\{y_1...y_{|\mathcal{Y}|}\}$, not the synonyms of labels $\tilde{y}_1...\tilde{y}_{|\mathcal{Y}|}$.

\section{Experimental Setup}\label{sec:exp-setup}\subsection{Data}\label{subsec:data}

\paragraph{Text corpus.}
We use the Demix corpus from \citet{suchin2021demix}, a raw text corpus 
that is not designated for any downstream task.
It consists of 16 diverse domains, including Wikipedia, news, Amazon reviews, Yelp reviews, Twitter, and more, all in English.
A full list is provided in Table~\ref{tab:corpuses} in Appendix~\ref{app:statistics}.
We subsample up to 10M paragraphs from each domain, and split each paragraph into sentences in order to perform a sentence-level retrieval.
More details are provided in Appendix~\ref{app:statistics}.

\begin{table}[t!]
    \centering \footnotesize
    \setlength{\tabcolsep}{2.5pt}
    \begin{tabular}{l cccc
    }
    \toprule
        Method & Demo & Corpus & Similar & No-Copy 
        \\
    \midrule
        No-demos        & - & &\\
        Random inputs   & pseudo & \checkmark&\\
        \naive        & pseudo & \checkmark & \checkmark &\\
        \textbf{\ours\ (Ours)}          & pseudo & \checkmark& \checkmark & \checkmark\\
        ICL-gold (Oracle)   & $k$-shot & &  & 
        \\
        ICL-random (Oracle) & $k$-shot & &  & 
        \\
    \bottomrule
    \end{tabular}\vspace{-.1em}
    \caption{Comparison between \ours\ and baselines.
    `{\em Demo}' indicates the type of the \demos, either the $k$-shot training data ($k$-shot) or constructed from a raw corpus only (pseudo).
    `{\em Corpus}' indicates whether an external corpus is used.
    `{\em Similar}' indicates whether a similarity function is used.
    `{\em No-Copy}' indicates whether the method is designed to reduce the copying effect.
    }\label{tab:baselines}
\end{table}

\vspace{-.2em}
\paragraph{Evaluation datasets.}
We evaluate our methods on nine single-sentence classification datasets: CR~\citep{ding2008holistic}, Amz~\citep{zhang2015character}, Amz5~\citep{zhang2015character}, Yelp~\citep{zhang2015character}, Yelp5~\citep{zhang2015character}, Tweet-Eval~\citep{barbieri2020tweeteval}, MR~\citep{pang2004sentimental}, SST2~\citep{socher2013recursive} and SST5~\citep{socher2013recursive}.
Six out of the nine datasets are from domains that are represented in our corpus, while the other three (MR, SST2, and SST5) are not. 
This split allows us to measure domain coverage effects. Statistics are reported in Appendix~\ref{app:statistics}.

\subsection{Baselines}\label{subsec:baselines}
We compare \ours\ with the following zero-shot methods. See Table~\ref{tab:baselines} for their comparison.

\vspace{.3em}
\noindent
\textbf{No-demonstrations (No-demos)} predicts $\mathrm{argmax}_{y\in \mathcal{Y}}P(y\mid x)$ without using any demonstrations. This is a previously-used zero-shot method~\citep{radford2019language,brown2020language}.

\vspace{.3em}
\noindent
\textbf{Random inputs} selects $x_1...x_k$ from $\mathcal{C}$ uniformly at random, without considering the similarity score with $x$.
It then pairs each $x_i$ with a random label from $\mathcal{Y}$ and uses in-context learning as in Section~\ref{subsec:method-icl}.
This baseline uses \pdemos, but does not consider the similarity between the test input and the \pdemos.




\vspace{.3em}
\noindent
\textbf{\naive} is a version of \ours\ that uses the most naive retrieval method without the \pn\ adjustment (Section~\ref{subsec:method-retrieval}) or synonym labeling (Section~\ref{subsec:method-labeling}). 
This method encourages the {\em relevance} of the \pdemos\ the most, but does not reduce the copying effect.

\vspace{.3em}
We also compare with methods that use the training data, and call them {\em Oracle} baselines. 

\vspace{.3em}
\noindent
\textbf{ICL-gold (Oracle)} uses $k$ input-label pairs from the training data 
and in-context learning. This is equivalent to the standard in-context learning, first proposed by \citet{brown2020language}.

\vspace{.3em}
\noindent
\textbf{ICL-random (Oracle)} uses $k$ inputs from the training data and pairs each input with a random label sampled from $\mathcal{Y}$ uniformly at random, and uses in-context learning~\citep{min2022rethinking}.

\newcommand{\rt}[2]{$#1_{#2}$}

\newcommand{\rgamzp}{\rt{50.0}{0.0}}

\newcommand{\cjndamzp}{\rt{86.1}{0.0}}
\newcommand{\cjgamzp}{\rt{90.9}{0.9}}
\newcommand{\cjramzp}{\rt{91.3}{1.4}}
\newcommand{\cjrsamzp}{\rt{81.8}{3.2}}
\newcommand{\cjnoamzp}{\rt{81.6}{0.5}}
\newcommand{\cjpnoamzp}{\rt{84.8}{0.6}}
\newcommand{\cjnamzp}{\rt{88.6}{0.2}}
\newcommand{\cjdnamzp}{\rt{88.7}{0.5}}
\newcommand{\cjpnamzp}{\rt{\mathbf{88.9}}{\mathbf{0.2}}}
\newcommand{\cjdramzp}{\rt{0}{0}}

\newcommand{\djndamzp}{\rt{87.3}{0.0}}
\newcommand{\djgamzp}{\rt{95.8}{0.1}}
\newcommand{\djramzp}{\rt{87.8}{7.5}}
\newcommand{\djrsamzp}{\rt{91.2}{2.8}}
\newcommand{\djnoamzp}{\rt{89.3}{0.6}}
\newcommand{\djpnoamzp}{\rt{90.7}{0.2}}
\newcommand{\djnamzp}{\rt{94.0}{0.3}}
\newcommand{\djdnamzp}{\rt{94.4}{0.2}}
\newcommand{\djpnamzp}{\rt{\mathbf{94.9}}{\mathbf{0.1}}}
\newcommand{\djdramzp}{\rt{0}{0}}

\newcommand{\cnndamzp}{\rt{63.2}{0.0}}
\newcommand{\cngamzp}{\rt{90.3}{0.8}}
\newcommand{\cnramzp}{\rt{88.5}{1.5}}
\newcommand{\cnrsamzp}{\rt{70.4}{2.3}}
\newcommand{\cnnoamzp}{\rt{78.8}{0.9}}
\newcommand{\cnpnoamzp}{\rt{80.0}{1.1}}
\newcommand{\cnnamzp}{\rt{84.0}{0.3}}
\newcommand{\cndnamzp}{\rt{\mathbf{84.5}}{\mathbf{0.9}}}
\newcommand{\cnpnamzp}{\rt{\mathbf{84.3}}{\mathbf{0.7}}}
\newcommand{\cndramzp}{\rt{82.0}{0.4}}

\newcommand{\dnndamzp}{\rt{50.8}{0.0}}
\newcommand{\dngamzp}{\rt{95.6}{0.5}}
\newcommand{\dnramzp}{\rt{92.9}{2.5}}
\newcommand{\dnrsamzp}{\rt{83.5}{12.9}}
\newcommand{\dnnoamzp}{\rt{87.5}{0.7}}
\newcommand{\dnpnoamzp}{\rt{87.6}{0.7}}
\newcommand{\dnnamzp}{\rt{\mathbf{94.9}}{\mathbf{0.4}}}
\newcommand{\dndnamzp}{\rt{94.8}{0.3}}
\newcommand{\dnpnamzp}{\rt{\mathbf{94.0}}{\mathbf{0.1}}}
\newcommand{\dndramzp}{\rt{94.1}{0.5}}

\newcommand{\rgcr}{\rt{50.0}{0.0}}

\newcommand{\cjndcr}{\rt{73.2}{0.0}}
\newcommand{\cjgcr}{\rt{84.4}{2.8}}
\newcommand{\cjrcr}{\rt{82.3}{1.3}}
\newcommand{\cjrscr}{\rt{77.8}{2.4}}
\newcommand{\cjnocr}{\rt{62.1}{0.8}}
\newcommand{\cjpnocr}{\rt{79.7}{0.5}}
\newcommand{\cjncr}{\rt{73.2}{0.3}}
\newcommand{\cjdncr}{\rt{77.6}{0.6}}
\newcommand{\cjpncr}{\rt{\mathbf{80.1}}{\mathbf{0.1}}}
\newcommand{\cjdrcr}{\rt{\mathbf{0}}{\mathbf{0}}}

\newcommand{\djndcr}{\rt{50.6}{0.0}}
\newcommand{\djgcr}{\rt{68.7}{13.9}}
\newcommand{\djrcr}{\rt{79.1}{10.0}}
\newcommand{\djrscr}{\rt{71.1}{15.0}}
\newcommand{\djnocr}{\rt{65.2}{0.9}}
\newcommand{\djpnocr}{\rt{70.5}{0.5}}
\newcommand{\djncr}{\rt{73.1}{0.5}}
\newcommand{\djdncr}{\rt{77.2}{0.4}}
\newcommand{\djpncr}{\rt{\mathbf{78.8}}{\mathbf{0.4}}}
\newcommand{\djdrcr}{\rt{\mathbf{0}}{\mathbf{0}}}

\newcommand{\cnndcr}{\rt{57.2}{0.0}}
\newcommand{\cngcr}{\rt{85.5}{2.3}}
\newcommand{\cnrcr}{\rt{78.1}{3.3}}
\newcommand{\cnrscr}{\rt{68.0}{4.2}}
\newcommand{\cnnocr}{\rt{62.4}{0.2}}
\newcommand{\cnpnocr}{\rt{75.4}{0.4}}
\newcommand{\cnncr}{\rt{72.0}{0.5}}
\newcommand{\cndncr}{\rt{75.9}{0.4}}
\newcommand{\cnpncr}{\rt{\mathbf{79.0}}{\mathbf{0.2}}}
\newcommand{\cndrcr}{\rt{\mathbf{80.9}}{\mathbf{0.9}}}

\newcommand{\dnndcr}{\rt{61.5}{0.0}}
\newcommand{\dngcr}{\rt{78.5}{14.8}}
\newcommand{\dnrcr}{\rt{78.5}{13.6}}
\newcommand{\dnrscr}{\rt{72.5}{13.7}}
\newcommand{\dnnocr}{\rt{76.2}{0.3}}
\newcommand{\dnpnocr}{\rt{82.2}{0.3}}
\newcommand{\dnncr}{\rt{88.6}{0.3}}
\newcommand{\dndncr}{\rt{90.9}{0.3}}
\newcommand{\dnpncr}{\rt{\mathbf{91.4}}{\mathbf{0.3}}}
\newcommand{\dndrcr}{\rt{90.8}{0.3}}

\newcommand{\rgmr}{\rt{50.0}{0.0}}

\newcommand{\cjndmr}{\rt{65.7}{0.0}}
\newcommand{\cjgmr}{\rt{86.9}{0.2}}
\newcommand{\cjrmr}{\rt{86.6}{0.3}}
\newcommand{\cjrsmr}{\rt{76.2}{3.6}}
\newcommand{\cjnomr}{\rt{68.8}{0.4}}
\newcommand{\cjpnomr}{\rt{77.7}{0.7}}
\newcommand{\cjnmr}{\rt{78.6}{0.4}}
\newcommand{\cjdnmr}{\rt{79.4}{0.2}}
\newcommand{\cjpnmr}{\rt{\mathbf{81.9}}{\mathbf{0.1}}}
\newcommand{\cjdrmr}{\rt{0}{0}}

\newcommand{\djndmr}{\rt{51.7}{0.0}}
\newcommand{\djgmr}{\rt{84.0}{6.8}}
\newcommand{\djrmr}{\rt{87.3}{3.6}}
\newcommand{\djrsmr}{\rt{68.2}{12.1}}
\newcommand{\djnomr}{\rt{64.6}{0.4}}
\newcommand{\djpnomr}{\rt{69.5}{0.8}}
\newcommand{\djnmr}{\rt{75.4}{0.1}}
\newcommand{\djdnmr}{\rt{78.6}{0.3}}
\newcommand{\djpnmr}{\rt{\mathbf{81.0}}{\mathbf{0.3}}}
\newcommand{\djdrmr}{\rt{0}{0}}

\newcommand{\cnndmr}{\rt{58.7}{0.0}}
\newcommand{\cngmr}{\rt{86.2}{0.8}}
\newcommand{\cnrmr}{\rt{86.3}{0.9}}
\newcommand{\cnrsmr}{\rt{65.0}{4.9}}
\newcommand{\cnnomr}{\rt{63.5}{0.8}}
\newcommand{\cnpnomr}{\rt{67.8}{0.5}}
\newcommand{\cnnmr}{\rt{69.9}{0.5}}
\newcommand{\cndnmr}{\rt{71.9}{0.8}}
\newcommand{\cnpnmr}{\rt{\mathbf{73.2}}{\mathbf{0.3}}}
\newcommand{\cndrmr}{\rt{69.3}{0.4}}

\newcommand{\dnndmr}{\rt{49.9}{0.0}}
\newcommand{\dngmr}{\rt{89.0}{0.9}}
\newcommand{\dnrmr}{\rt{81.2}{13.7}}
\newcommand{\dnrsmr}{\rt{74.9}{8.7}}
\newcommand{\dnnomr}{\rt{71.7}{1.1}}
\newcommand{\dnpnomr}{\rt{74.1}{0.3}}
\newcommand{\dnnmr}{\rt{82.2}{0.4}}
\newcommand{\dndnmr}{\rt{\mathbf{84.2}}{\mathbf{0.5}}}
\newcommand{\dnpnmr}{\rt{\mathbf{84.0}}{\mathbf{0.4}}}
\newcommand{\dndrmr}{\rt{81.9}{0.2}}

\newcommand{\rgsstt}{\rt{50.0}{0.0}}

\newcommand{\cjndsstt}{\rt{66.3}{0.0}}
\newcommand{\cjgsstt}{\rt{88.8}{1.3}}
\newcommand{\cjrsstt}{\rt{86.1}{2.1}}
\newcommand{\cjrssstt}{\rt{78.6}{3.6}}
\newcommand{\cjnosstt}{\rt{67.8}{0.8}}
\newcommand{\cjpnosstt}{\rt{77.6}{0.7}}
\newcommand{\cjnsstt}{\rt{79.2}{0.5}}
\newcommand{\cjdnsstt}{\rt{82.3}{0.4}}
\newcommand{\cjpnsstt}{\rt{\mathbf{82.6}}{\mathbf{0.2}}}
\newcommand{\cjdrsstt}{\rt{0}{0}}

\newcommand{\djndsstt}{\rt{52.9}{0.0}}
\newcommand{\djgsstt}{\rt{91.1}{3.2}}
\newcommand{\djrsstt}{\rt{82.6}{9.7}}
\newcommand{\djrssstt}{\rt{69.9}{12.9}}
\newcommand{\djnosstt}{\rt{66.1}{0.0}}
\newcommand{\djpnosstt}{\rt{70.6}{0.9}}
\newcommand{\djnsstt}{\rt{76.6}{0.9}}
\newcommand{\djdnsstt}{\rt{79.5}{0.7}}
\newcommand{\djpnsstt}{\rt{\mathbf{82.6}}{\mathbf{0.2}}}
\newcommand{\djdrsstt}{\rt{0}{0}}

\newcommand{\cnndsstt}{\rt{61.9}{0.0}}
\newcommand{\cngsstt}{\rt{89.4}{0.9}}
\newcommand{\cnrsstt}{\rt{88.1}{1.6}}
\newcommand{\cnrssstt}{\rt{66.4}{5.2}}
\newcommand{\cnnosstt}{\rt{62.8}{0.7}}
\newcommand{\cnpnosstt}{\rt{68.2}{0.7}}
\newcommand{\cnnsstt}{\rt{71.1}{0.3}}
\newcommand{\cndnsstt}{\rt{72.9}{1.6}}
\newcommand{\cnpnsstt}{\rt{\mathbf{74.3}}{\mathbf{0.2}}}
\newcommand{\cndrsstt}{\rt{72.1}{1.4}}

\newcommand{\dnndsstt}{\rt{49.1}{0.0}}
\newcommand{\dngsstt}{\rt{88.6}{5.1}}
\newcommand{\dnrsstt}{\rt{76.9}{13.8}}
\newcommand{\dnrssstt}{\rt{78.2}{9.4}}
\newcommand{\dnnosstt}{\rt{73.8}{1.0}}
\newcommand{\dnpnosstt}{\rt{77.7}{0.9}}
\newcommand{\dnnsstt}{\rt{85.4}{0.3}}
\newcommand{\dndnsstt}{\rt{86.9}{0.5}}
\newcommand{\dnpnsstt}{\rt{\mathbf{87.8}}{\mathbf{0.7}}}
\newcommand{\dndrsstt}{\rt{84.7}{0.3}}

\newcommand{\rgteh}{\rt{38.1}{0.0}}

\newcommand{\cjndteh}{\rt{\mathbf{47.6}}{\mathbf{0.0}}}
\newcommand{\cjgteh}{\rt{48.0}{1.8}}
\newcommand{\cjrteh}{\rt{46.8}{2.6}}
\newcommand{\cjrsteh}{\rt{41.5}{1.1}}
\newcommand{\cjnoteh}{\rt{42.2}{1.0}}
\newcommand{\cjpnoteh}{\rt{44.1}{0.6}}
\newcommand{\cjnteh}{\rt{46.5}{0.2}}
\newcommand{\cjdnteh}{\rt{46.8}{0.6}}
\newcommand{\cjpnteh}{\rt{46.8}{0.5}}
\newcommand{\cjdrteh}{\rt{0}{0}}

\newcommand{\djndteh}{\rt{\mathbf{39.5}}{\mathbf{0.0}}}
\newcommand{\djgteh}{\rt{35.0}{5.1}}
\newcommand{\djrteh}{\rt{33.4}{2.7}}
\newcommand{\djrsteh}{\rt{28.8}{6.7}}
\newcommand{\djnoteh}{\rt{32.3}{0.4}}
\newcommand{\djpnoteh}{\rt{33.7}{0.4}}
\newcommand{\djnteh}{\rt{20.6}{0.2}}
\newcommand{\djdnteh}{\rt{20.7}{0.2}}
\newcommand{\djpnteh}{\rt{20.5}{0.1}}
\newcommand{\djdrteh}{\rt{0}{0}}

\newcommand{\cnndteh}{\rt{28.7}{0.0}}
\newcommand{\cngteh}{\rt{47.9}{1.9}}
\newcommand{\cnrteh}{\rt{44.0}{1.1}}
\newcommand{\cnrsteh}{\rt{34.6}{4.9}}
\newcommand{\cnnoteh}{\rt{38.9}{0.5}}
\newcommand{\cnpnoteh}{\rt{42.5}{0.8}}
\newcommand{\cnnteh}{\rt{45.1}{0.6}}
\newcommand{\cndnteh}{\rt{45.1}{0.6}}
\newcommand{\cnpnteh}{\rt{\mathbf{46.7}}{\mathbf{0.6}}}
\newcommand{\cndrteh}{\rt{45.0}{0.8}}

\newcommand{\dnndteh}{\rt{30.8}{0.0}}
\newcommand{\dngteh}{\rt{32.8}{6.5}}
\newcommand{\dnrteh}{\rt{33.1}{3.9}}
\newcommand{\dnrsteh}{\rt{36.4}{9.5}}
\newcommand{\dnnoteh}{\rt{\mathbf{40.2}}{\mathbf{0.9}}}
\newcommand{\dnpnoteh}{\rt{38.6}{0.7}}
\newcommand{\dnnteh}{\rt{33.2}{1.0}}
\newcommand{\dndnteh}{\rt{34.4}{0.7}}
\newcommand{\dnpnteh}{\rt{35.2}{0.9}}
\newcommand{\dndrteh}{\rt{33.2}{1.2}}

\newcommand{\rgyelpp}{\rt{50.0}{0.0}}

\newcommand{\cjndyelpp}{\rt{88.0}{0.0}}
\newcommand{\cjgyelpp}{\rt{91.0}{0.1}}
\newcommand{\cjryelpp}{\rt{91.1}{0.3}}
\newcommand{\cjrsyelpp}{\rt{84.2}{4.6}}
\newcommand{\cjnoyelpp}{\rt{81.4}{0.3}}
\newcommand{\cjpnoyelpp}{\rt{86.4}{0.4}}
\newcommand{\cjnyelpp}{\rt{87.0}{0.1}}
\newcommand{\cjdnyelpp}{\rt{87.9}{0.1}}
\newcommand{\cjpnyelpp}{\rt{\mathbf{88.4}}{\mathbf{0.1}}}
\newcommand{\cjdryelpp}{\rt{0}{0}}

\newcommand{\djndyelpp}{\rt{92.3}{0.0}}
\newcommand{\djgyelpp}{\rt{96.4}{0.4}}
\newcommand{\djryelpp}{\rt{94.5}{1.9}}
\newcommand{\djrsyelpp}{\rt{91.5}{3.5}}
\newcommand{\djnoyelpp}{\rt{91.7}{0.6}}
\newcommand{\djpnoyelpp}{\rt{92.4}{0.5}}
\newcommand{\djnyelpp}{\rt{95.3}{0.2}}
\newcommand{\djdnyelpp}{\rt{95.9}{0.2}}
\newcommand{\djpnyelpp}{\rt{\mathbf{96.0}}{\mathbf{0.1}}}
\newcommand{\djdryelpp}{\rt{0}{0}}

\newcommand{\cnndyelpp}{\rt{57.0}{0.0}}
\newcommand{\cngyelpp}{\rt{86.8}{2.8}}
\newcommand{\cnryelpp}{\rt{88.0}{1.7}}
\newcommand{\cnrsyelpp}{\rt{73.0}{3.1}}
\newcommand{\cnnoyelpp}{\rt{79.1}{0.8}}
\newcommand{\cnpnoyelpp}{\rt{82.3}{0.3}}
\newcommand{\cnnyelpp}{\rt{85.5}{0.2}}
\newcommand{\cndnyelpp}{\rt{\mathbf{87.3}}{\mathbf{0.6}}}
\newcommand{\cnpnyelpp}{\rt{\mathbf{87.0}}{\mathbf{0.4}}}
\newcommand{\cndryelpp}{\rt{87.2}{0.3}}

\newcommand{\dnndyelpp}{\rt{72.2}{0.0}}
\newcommand{\dngyelpp}{\rt{91.7}{3.6}}
\newcommand{\dnryelpp}{\rt{88.5}{4.3}}
\newcommand{\dnrsyelpp}{\rt{85.0}{8.4}}
\newcommand{\dnnoyelpp}{\rt{89.0}{0.8}}
\newcommand{\dnpnoyelpp}{\rt{89.3}{0.4}}
\newcommand{\dnnyelpp}{\rt{\mathbf{92.8}}{\mathbf{0.4}}}
\newcommand{\dndnyelpp}{\rt{92.7}{0.3}}
\newcommand{\dnpnyelpp}{\rt{\mathbf{92.2}}{\mathbf{0.3}}}
\newcommand{\dndryelpp}{\rt{92.5}{0.2}}

\newcommand{\rgamaz}{\rt{20.0}{0.0}}

\newcommand{\cjndamaz}{\rt{34.4}{0.0}}
\newcommand{\cjgamaz}{\rt{45.5}{3.2}}
\newcommand{\cjramaz}{\rt{44.9}{2.0}}
\newcommand{\cjrsamaz}{\rt{38.1}{1.6}}
\newcommand{\cjnoamaz}{\rt{41.7}{0.4}}
\newcommand{\cjpnoamaz}{\rt{42.4}{0.8}}
\newcommand{\cjnamaz}{\rt{45.2}{0.6}}
\newcommand{\cjdnamaz}{\rt{45.1}{0.4}}
\newcommand{\cjpnamaz}{\rt{\mathbf{46.5}}{\mathbf{0.4}}}
\newcommand{\cjdramaz}{\rt{0}{0}}

\newcommand{\djndamaz}{\rt{30.4}{0.0}}
\newcommand{\djgamaz}{\rt{49.0}{3.8}}
\newcommand{\djramaz}{\rt{41.1}{4.8}}
\newcommand{\djrsamaz}{\rt{37.5}{5.2}}
\newcommand{\djnoamaz}{\rt{\mathbf{39.6}}{\mathbf{0.4}}}
\newcommand{\djpnoamaz}{\rt{\mathbf{40.2}}{\mathbf{0.8}}}
\newcommand{\djnamaz}{\rt{37.8}{0.4}}
\newcommand{\djdnamaz}{\rt{38.1}{0.3}}
\newcommand{\djpnamaz}{\rt{38.5}{0.3}}
\newcommand{\djdramaz}{\rt{0}{0}}

\newcommand{\cnndamaz}{\rt{27.5}{0.0}}
\newcommand{\cngamaz}{\rt{41.6}{1.8}}
\newcommand{\cnramaz}{\rt{39.8}{1.4}}
\newcommand{\cnrsamaz}{\rt{27.9}{1.9}}
\newcommand{\cnnoamaz}{\rt{34.7}{1.2}}
\newcommand{\cnpnoamaz}{\rt{35.1}{1.0}}
\newcommand{\cnnamaz}{\rt{37.3}{0.4}}
\newcommand{\cndnamaz}{\rt{\mathbf{38.5}}{\mathbf{0.4}}}
\newcommand{\cnpnamaz}{\rt{\mathbf{37.8}}{\mathbf{0.5}}}
\newcommand{\cndramaz}{\rt{36.0}{0.9}}

\newcommand{\dnndamaz}{\rt{20.2}{0.0}}
\newcommand{\dngamaz}{\rt{47.0}{2.7}}
\newcommand{\dnramaz}{\rt{45.6}{1.6}}
\newcommand{\dnrsamaz}{\rt{38.7}{3.6}}
\newcommand{\dnnoamaz}{\rt{41.2}{0.9}}
\newcommand{\dnpnoamaz}{\rt{42.0}{1.1}}
\newcommand{\dnnamaz}{\rt{\mathbf{41.5}}{\mathbf{0.6}}}
\newcommand{\dndnamaz}{\rt{41.3}{0.6}}
\newcommand{\dnpnamaz}{\rt{\mathbf{41.2}}{\mathbf{0.4}}}
\newcommand{\dndramaz}{\rt{39.1}{0.3}}

\newcommand{\rgsstf}{\rt{21.5}{0.0}}

\newcommand{\cjndsstf}{\rt{21.9}{0.0}}
\newcommand{\cjgsstf}{\rt{42.1}{1.1}}
\newcommand{\cjrsstf}{\rt{41.8}{0.9}}
\newcommand{\cjrssstf}{\rt{33.9}{3.6}}
\newcommand{\cjnosstf}{\rt{32.4}{0.6}}
\newcommand{\cjpnosstf}{\rt{34.9}{0.5}}
\newcommand{\cjnsstf}{\rt{35.9}{0.5}}
\newcommand{\cjdnsstf}{\rt{38.0}{0.8}}
\newcommand{\cjpnsstf}{\rt{\mathbf{38.7}}{\mathbf{0.5}}}
\newcommand{\cjdrsstf}{\rt{0}{0}}

\newcommand{\djndsstf}{\rt{26.8}{0.0}}
\newcommand{\djgsstf}{\rt{42.9}{0.9}}
\newcommand{\djrsstf}{\rt{35.9}{3.5}}
\newcommand{\djrssstf}{\rt{30.1}{8.2}}
\newcommand{\djnosstf}{\rt{30.9}{0.6}}
\newcommand{\djpnosstf}{\rt{\mathbf{32.9}}{\mathbf{0.6}}}
\newcommand{\djnsstf}{\rt{28.9}{0.4}}
\newcommand{\djdnsstf}{\rt{30.1}{0.4}}
\newcommand{\djpnsstf}{\rt{\mathbf{30.9}}{\mathbf{0.3}}}
\newcommand{\djdrsstf}{\rt{28.5}{0.8}}

\newcommand{\cnndsstf}{\rt{23.8}{0.0}}
\newcommand{\cngsstf}{\rt{40.8}{1.1}}
\newcommand{\cnrsstf}{\rt{39.9}{1.2}}
\newcommand{\cnrssstf}{\rt{26.8}{3.6}}
\newcommand{\cnnosstf}{\rt{29.9}{0.8}}
\newcommand{\cnpnosstf}{\rt{31.0}{0.6}}
\newcommand{\cnnsstf}{\rt{32.0}{0.2}}
\newcommand{\cndnsstf}{\rt{31.9}{0.5}}
\newcommand{\cnpnsstf}{\rt{\mathbf{33.2}}{\mathbf{0.3}}}
\newcommand{\cndrsstf}{\rt{30.8}{0.6}}

\newcommand{\dnndsstf}{\rt{17.5}{0.0}}
\newcommand{\dngsstf}{\rt{43.0}{3.1}}
\newcommand{\dnrsstf}{\rt{37.5}{3.1}}
\newcommand{\dnrssstf}{\rt{37.5}{6.2}}
\newcommand{\dnnosstf}{\rt{\mathbf{34.0}}{\mathbf{0.5}}}
\newcommand{\dnpnosstf}{\rt{\mathbf{35.6}}{\mathbf{0.6}}}
\newcommand{\dnnsstf}{\rt{33.0}{0.1}}
\newcommand{\dndnsstf}{\rt{34.3}{0.8}}
\newcommand{\dnpnsstf}{\rt{33.3}{0.6}}
\newcommand{\dndrsstf}{\rt{29.7}{0.5}}

\newcommand{\rgyelpf}{\rt{20.0}{0.0}}

\newcommand{\cjndyelpf}{\rt{36.6}{0.0}}
\newcommand{\cjgyelpf}{\rt{47.4}{1.3}}
\newcommand{\cjryelpf}{\rt{48.0}{1.5}}
\newcommand{\cjrsyelpf}{\rt{40.5}{1.4}}
\newcommand{\cjnoyelpf}{\rt{41.8}{0.8}}
\newcommand{\cjpnoyelpf}{\rt{\mathbf{43.6}}{\mathbf{0.5}}}
\newcommand{\cjnyelpf}{\rt{44.3}{0.5}}
\newcommand{\cjdnyelpf}{\rt{\mathbf{44.9}}{\mathbf{0.6}}}
\newcommand{\cjpnyelpf}{\rt{\mathbf{44.2}}{\mathbf{0.3}}}
\newcommand{\cjdryelpf}{\rt{0}{0}}

\newcommand{\djndyelpf}{\rt{28.7}{0.0}}
\newcommand{\djgyelpf}{\rt{47.5}{5.8}}
\newcommand{\djryelpf}{\rt{43.5}{3.5}}
\newcommand{\djrsyelpf}{\rt{36.4}{6.1}}
\newcommand{\djnoyelpf}{\rt{\mathbf{41.2}}{\mathbf{0.8}}}
\newcommand{\djpnoyelpf}{\rt{\mathbf{42.5}}{\mathbf{0.5}}}
\newcommand{\djnyelpf}{\rt{39.6}{0.4}}
\newcommand{\djdnyelpf}{\rt{40.3}{0.4}}
\newcommand{\djpnyelpf}{\rt{40.8}{0.3}}
\newcommand{\djdryelpf}{\rt{41.9}{0.7}}

\newcommand{\cnndyelpf}{\rt{28.6}{0.0}}
\newcommand{\cngyelpf}{\rt{43.5}{0.7}}
\newcommand{\cnryelpf}{\rt{43.5}{1.6}}
\newcommand{\cnrsyelpf}{\rt{29.1}{1.9}}
\newcommand{\cnnoyelpf}{\rt{36.9}{0.8}}
\newcommand{\cnpnoyelpf}{\rt{38.0}{1.0}}
\newcommand{\cnnyelpf}{\rt{39.8}{0.9}}
\newcommand{\cndnyelpf}{\rt{\mathbf{40.4}}{\mathbf{0.5}}}
\newcommand{\cnpnyelpf}{\rt{\mathbf{39.9}}{\mathbf{1.0}}}
\newcommand{\cndryelpf}{\rt{40.4}{0.7}}

\newcommand{\dnndyelpf}{\rt{21.3}{0.0}}
\newcommand{\dngyelpf}{\rt{40.6}{3.1}}
\newcommand{\dnryelpf}{\rt{41.3}{3.5}}
\newcommand{\dnrsyelpf}{\rt{37.1}{2.6}}
\newcommand{\dnnoyelpf}{\rt{\mathbf{39.1}}{\mathbf{0.6}}}
\newcommand{\dnpnoyelpf}{\rt{\mathbf{39.2}}{\mathbf{0.8}}}
\newcommand{\dnnyelpf}{\rt{38.9}{0.4}}
\newcommand{\dndnyelpf}{\rt{\mathbf{39.1}}{\mathbf{0.6}}}
\newcommand{\dnpnyelpf}{\rt{38.6}{0.3}}
\newcommand{\dndryelpf}{\rt{37.7}{0.5}}

\newcommand{\rgiave}{\rt{38.0}{0.0}}

\newcommand{\cjndiave}{\rt{61.0}{0.0}}
\newcommand{\cjgiave}{\rt{67.9}{1.7}}
\newcommand{\cjriave}{\rt{67.4}{1.5}}
\newcommand{\cjrsiave}{\rt{60.7}{2.4}}
\newcommand{\cjnoiave}{\rt{58.5}{0.6}}
\newcommand{\cjpnoiave}{\rt{63.5}{0.6}}
\newcommand{\cjniave}{\rt{64.1}{0.3}}
\newcommand{\cjdniave}{\rt{65.2}{0.5}}
\newcommand{\cjpniave}{\rt{\mathbf{65.8}}{\mathbf{0.3}}}
\newcommand{\cjdriave}{\rt{0}{0}}

\newcommand{\djndiave}{\rt{54.8}{0.0}}
\newcommand{\djgiave}{\rt{65.4}{4.9}}
\newcommand{\djriave}{\rt{63.2}{5.1}}
\newcommand{\djrsiave}{\rt{59.4}{6.6}}
\newcommand{\djnoiave}{\rt{59.9}{0.6}}
\newcommand{\djpnoiave}{\rt{61.7}{0.5}}
\newcommand{\djniave}{\rt{60.1}{0.3}}
\newcommand{\djdniave}{\rt{61.1}{0.3}}
\newcommand{\djpniave}{\rt{\mathbf{61.6}}{\mathbf{0.3}}}
\newcommand{\djdriave}{\rt{0}{0}}

\newcommand{\cnndiave}{\rt{43.7}{0.0}}
\newcommand{\cngiave}{\rt{65.9}{1.7}}
\newcommand{\cnriave}{\rt{63.7}{1.8}}
\newcommand{\cnrsiave}{\rt{50.5}{3.1}}
\newcommand{\cnnoiave}{\rt{55.1}{0.7}}
\newcommand{\cnpnoiave}{\rt{58.9}{0.8}}
\newcommand{\cnniave}{\rt{60.6}{0.5}}
\newcommand{\cndniave}{\rt{62.0}{0.6}}
\newcommand{\cnpniave}{\rt{\mathbf{62.5}}{\mathbf{0.6}}}
\newcommand{\cndriave}{\rt{61.9}{0.7}}

\newcommand{\dnndiave}{\rt{42.8}{0.0}}
\newcommand{\dngiave}{\rt{64.4}{5.2}}
\newcommand{\dnriave}{\rt{63.3}{4.9}}
\newcommand{\dnrsiave}{\rt{58.9}{8.5}}
\newcommand{\dnnoiave}{\rt{62.2}{0.7}}
\newcommand{\dnpnoiave}{\rt{63.0}{0.7}}
\newcommand{\dnniave}{\rt{65.0}{0.5}}
\newcommand{\dndniave}{\rt{\mathbf{65.5}}{\mathbf{0.5}}}
\newcommand{\dnpniave}{\rt{\mathbf{65.4}}{\mathbf{0.4}}}
\newcommand{\dndriave}{\rt{64.6}{0.5}}

\newcommand{\rgoave}{\rt{40.5}{0.0}}

\newcommand{\cjndoave}{\rt{51.3}{0.0}}
\newcommand{\cjgoave}{\rt{72.6}{0.9}}
\newcommand{\cjroave}{\rt{71.5}{1.1}}
\newcommand{\cjrsoave}{\rt{62.9}{3.6}}
\newcommand{\cjnooave}{\rt{56.3}{0.6}}
\newcommand{\cjpnooave}{\rt{63.4}{0.6}}
\newcommand{\cjnoave}{\rt{64.6}{0.5}}
\newcommand{\cjdnoave}{\rt{66.6}{0.5}}
\newcommand{\cjpnoave}{\rt{\mathbf{67.7}}{\mathbf{0.3}}}
\newcommand{\cjdroave}{\rt{0}{0}}

\newcommand{\djndoave}{\rt{43.8}{0.0}}
\newcommand{\djgoave}{\rt{72.7}{4.0}}
\newcommand{\djroave}{\rt{68.6}{5.6}}
\newcommand{\djrsoave}{\rt{56.1}{11.1}}
\newcommand{\djnooave}{\rt{53.9}{0.3}}
\newcommand{\djpnooave}{\rt{57.7}{0.8}}
\newcommand{\djnoave}{\rt{60.3}{0.5}}
\newcommand{\djdnoave}{\rt{62.7}{0.5}}
\newcommand{\djpnoave}{\rt{\mathbf{64.8}}{\mathbf{0.3}}}
\newcommand{\djdroave}{\rt{0}{0}}

\newcommand{\cnndoave}{\rt{48.1}{0.0}}
\newcommand{\cngoave}{\rt{72.1}{0.9}}
\newcommand{\cnroave}{\rt{71.4}{1.2}}
\newcommand{\cnrsoave}{\rt{52.7}{4.6}}
\newcommand{\cnnooave}{\rt{55.1}{0.7}}
\newcommand{\cnpnooave}{\rt{55.1}{0.7}}
\newcommand{\cnnoave}{\rt{57.7}{0.3}}
\newcommand{\cndnoave}{\rt{58.9}{1.0}}
\newcommand{\cnpnoave}{\rt{\mathbf{60.2}}{\mathbf{0.3}}}
\newcommand{\cndroave}{\rt{57.4}{0.8}}

\newcommand{\dnndoave}{\rt{38.8}{0.0}}
\newcommand{\dngoave}{\rt{73.5}{3.0}}
\newcommand{\dnroave}{\rt{65.2}{10.2}}
\newcommand{\dnrsoave}{\rt{63.5}{8.1}}
\newcommand{\dnnooave}{\rt{59.8}{0.9}}
\newcommand{\dnpnooave}{\rt{59.8}{0.9}}
\newcommand{\dnnoave}{\rt{66.9}{0.3}}
\newcommand{\dndnoave}{\rt{\mathbf{68.5}}{\mathbf{0.6}}}
\newcommand{\dnpnoave}{\rt{\mathbf{68.4}}{\mathbf{0.6}}}
\newcommand{\dndroave}{\rt{65.4}{0.3}}

\newcommand{\ctndamzp}{\rt{77.2}{0.0}}
\newcommand{\ctgamzp}{\rt{86.0}{3.6}}
\newcommand{\ctramzp}{\rt{83.4}{4.8}}
\newcommand{\ctrsamzp}{\rt{80.0}{0.8}}
\newcommand{\ctdnamzp}{\rt{88.5}{0.1}}
\newcommand{\ctpnamzp}{\rt{\mathbf{89.1}}{\mathbf{0.3}}}

\newcommand{\dtndamzp}{\rt{88.2}{0.0}}
\newcommand{\dtgamzp}{\rt{97.0}{0.2}}
\newcommand{\dtramzp}{\rt{95.4}{0.6}}
\newcommand{\dtrsamzp}{\rt{\mathbf{97.3}}{\mathbf{0.1}}}
\newcommand{\dtdnamzp}{\rt{\mathbf{93.1}}{\mathbf{0.7}}}
\newcommand{\dtpnamzp}{\rt{\mathbf{93.0}}{\mathbf{0.2}}}

\newcommand{\ctndcr}{\rt{76.6}{0.0}}
\newcommand{\ctgcr}{\rt{74.2}{7.4}}
\newcommand{\ctrcr}{\rt{73.9}{3.9}}
\newcommand{\ctrscr}{\rt{75.0}{1.2}}
\newcommand{\ctdncr}{\rt{76.0}{1.4}}
\newcommand{\ctpncr}{\rt{\mathbf{80.8}}{\mathbf{0.6}}}

\newcommand{\dtndcr}{\rt{68.4}{0.0}}
\newcommand{\dtgcr}{\rt{79.5}{9.5}}
\newcommand{\dtrcr}{\rt{81.0}{6.8}}
\newcommand{\dtrscr}{\rt{66.6}{12.4}}
\newcommand{\dtdncr}{\rt{67.4}{0.0}}
\newcommand{\dtpncr}{\rt{\mathbf{71.9}}{\mathbf{0.1}}}

\newcommand{\ctndteh}{\rt{36.2}{0.0}}
\newcommand{\ctgteh}{\rt{43.8}{0.2}}
\newcommand{\ctrteh}{\rt{41.4}{2.0}}
\newcommand{\ctrsteh}{\rt{37.7}{0.0}}
\newcommand{\ctdnteh}{\rt{\mathbf{42.4}}{\mathbf{0.6}}}
\newcommand{\ctpnteh}{\rt{\mathbf{41.4}}{\mathbf{0.4}}}

\newcommand{\dtndteh}{\rt{\mathbf{37.8}}{\mathbf{0.0}}}
\newcommand{\dtgteh}{\rt{30.5}{8.0}}
\newcommand{\dtrteh}{\rt{42.2}{39.4}}
\newcommand{\dtrsteh}{\rt{\mathbf{35.1}}{\mathbf{1.2}}}
\newcommand{\dtdnteh}{\rt{26.5}{0.8}}
\newcommand{\dtpnteh}{\rt{28.3}{0.4}}

\newcommand{\ctndyelpp}{\rt{\mathbf{88.0}}{\mathbf{0.0}}}
\newcommand{\ctgyelpp}{\rt{91.7}{0.9}}
\newcommand{\ctryelpp}{\rt{90.4}{1.4}}
\newcommand{\ctrsyelpp}{\rt{78.5}{3.1}}
\newcommand{\ctdnyelpp}{\rt{86.7}{0.3}}
\newcommand{\ctpnyelpp}{\rt{87.6}{0.0}}

\newcommand{\dtndyelpp}{\rt{96.4}{0.0}}
\newcommand{\dtgyelpp}{\rt{98.5}{0.1}}
\newcommand{\dtryelpp}{\rt{93.7}{2.1}}
\newcommand{\dtrsyelpp}{\rt{97.7}{0.5}}
\newcommand{\dtdnyelpp}{\rt{97.3}{0.1}}
\newcommand{\dtpnyelpp}{\rt{\mathbf{97.7}}{\mathbf{0.3}}}

\newcommand{\ctndsstt}{\rt{80.8}{0.0}}
\newcommand{\ctgsstt}{\rt{88.1}{1.1}}
\newcommand{\ctrsstt}{\rt{84.8}{1.2}}
\newcommand{\ctrssstt}{\rt{66.9}{2.3}}
\newcommand{\ctdnsstt}{\rt{81.6}{0.8}}
\newcommand{\ctpnsstt}{\rt{\mathbf{82.4}}{\mathbf{74.7}}}

\newcommand{\dtndsstt}{\rt{73.2}{0.0}}
\newcommand{\dtgsstt}{\rt{94.2}{0.2}}
\newcommand{\dtrsstt}{\rt{93.9}{0.5}}
\newcommand{\dtrssstt}{\rt{72.7}{19.1}}
\newcommand{\dtdnsstt}{\rt{76.1}{0.3}}
\newcommand{\dtpnsstt}{\rt{\mathbf{78.1}}{\mathbf{0.1}}}

\newcommand{\rgfave}{\rt{47.6}{0.0}}

\newcommand{\ctndave}{\rt{69.5}{0.0}}
\newcommand{\ctgave}{\rt{73.9}{3.0}}
\newcommand{\ctrave}{\rt{72.3}{3.0}}
\newcommand{\ctrsave}{\rt{67.8}{3.0}}
\newcommand{\ctdnave}{\rt{67.8}{1.3}}
\newcommand{\ctpnave}{\rt{\mathbf{73.4}}{\mathbf{0.6}}}

\newcommand{\dtndave}{\rt{72.7}{0.0}}
\newcommand{\dtgave}{\rt{79.3}{2.5}}
\newcommand{\dtrave}{\rt{77.4}{2.7}}
\newcommand{\dtrsave}{\rt{\mathbf{74.2}}{\mathbf{3.6}}}
\newcommand{\dtdnave}{\rt{71.1}{0.4}}
\newcommand{\dtpnave}{\rt{\mathbf{72.7}}{0.3}}

\definecolor{Gray}{gray}{0.85}
\newcolumntype{a}{>{\columncolor{Gray}}r}

\begin{table*}[ht!]
    \centering \scriptsize
    \setlength{\tabcolsep}{3pt}
    \begin{tabular}{l rrrrrra  rrra }
    \toprule
        \multirow{2}{*}{Method} & \multicolumn{7}{c}{Covered by $\mathcal{C}$} & \multicolumn{4}{c}{Not covered by $\mathcal{C}$} \\
        \cmidrule(lr){2-8} \cmidrule(lr){9-12}
        & CR & Amz & Amz5 & Yelp & Yelp5 & Tweet & Avg & MR & SST2 & SST5 & Avg \\
    \midrule
        Majority & \rgcr  & \rgamzp & \rgamaz & \rgyelpp & \rgyelpf & \rgteh & \rgiave & \rgmr & \rgsstt & \rgsstf & \rgoave \\
    \midrule
        \multicolumn{11}{l}{\textbf{\em{Channel GPT-J}}} \\
        No-demos & \cjndcr  & \cjndamzp & \cjndamaz & \cjndyelpp & \cjndyelpf & \cjndteh & \cjndiave & \cjndmr & \cjndsstt & \cjndsstf & \cjndoave \\
        Random inputs         & \cjrscr  & \cjrsamzp & \cjrsamaz & \cjrsyelpp & \cjrsyelpf & \cjrsteh & \cjrsiave & \cjrsmr & \cjrssstt & \cjrssstf & \cjrsoave \\
        
        \naive  & \cjnocr  & \cjnoamzp & \cjnoamaz & \cjnoyelpp & \cjnoyelpf & \cjnoteh & \cjnoiave & \cjnomr & \cjnosstt & \cjnosstf & \cjnooave \\
        \ours\ (Ours)  & \cjpncr  & \cjpnamzp & \cjpnamaz & \cjpnyelpp & \cjpnyelpf & \cjpnteh & \cjpniave & \cjpnmr & \cjpnsstt & \cjpnsstf & \cjpnoave \\
        \cmidrule(lr){1-12}
        ICL-gold (Oracle) & \cjgcr  & \cjgamzp & \cjgamaz & \cjgyelpp & \cjgyelpf & \cjgteh & \cjgiave & \cjgmr & \cjgsstt & \cjgsstf & \cjgoave \\
        ICL-random (Oracle)    & \cjrcr  & \cjramzp & \cjramaz & \cjryelpp & \cjryelpf & \cjrteh & \cjriave & \cjrmr & \cjrsstt & \cjrsstf & \cjroave \\
    \midrule
        \multicolumn{11}{l}{\textbf{\em{Direct GPT-J}}} \\
        No-demos & \djndcr  & \djndamzp & \djndamaz & \djndyelpp & \djndyelpf & \djndteh & \djndiave & \djndmr & \djndsstt & \djndsstf & \djndoave \\
        Random inputs         & \djrscr  & \djrsamzp & \djrsamaz & \djrsyelpp & \djrsyelpf & \djrsteh & \djrsiave & \djrsmr & \djrssstt & \djrssstf & \djrsoave \\
        \naive  & \djnocr  & \djnoamzp & \djnoamaz & \djnoyelpp & \djnoyelpf & \djnoteh & \djnoiave & \djnomr & \djnosstt & \djnosstf & \djnooave \\
        \ours\ (Ours)  & \djpncr  & \djpnamzp & \djpnamaz & \djpnyelpp & \djpnyelpf & \djpnteh & \djpniave & \djpnmr & \djpnsstt & \djpnsstf & \djpnoave \\
        \cmidrule(lr){1-12}
        ICL-gold (Oracle) & \djgcr  & \djgamzp & \djgamaz & \djgyelpp & \djgyelpf & \djgteh & \djgiave & \djgmr & \djgsstt & \djgsstf & \djgoave \\
        ICL-random (Oracle)    & \djrcr  & \djramzp & \djramaz & \djryelpp & \djryelpf & \djrteh & \djriave & \djrmr & \djrsstt & \djrsstf & \djroave \\
    \midrule
        \multicolumn{11}{l}{\textbf{\em{Channel GPT-NeoX}}} \\
        No-demos & \cnndcr  & \cnndamzp & \cnndamaz & \cnndyelpp & \cnndyelpf & \cnndteh & \cnndiave & \cnndmr & \cnndsstt & \cnndsstf & \cnndoave \\
        Random inputs         & \cnrscr  & \cnrsamzp & \cnrsamaz & \cnrsyelpp & \cnrsyelpf & \cnrsteh & \cnrsiave & \cnrsmr & \cnrssstt & \cnrssstf & \cnrsoave \\
        
        \naive  & \cnnocr  & \cnnoamzp & \cnnoamaz & \cnnoyelpp & \cnnoyelpf & \cnnoteh & \cnnoiave & \cnnomr & \cnnosstt & \cnnosstf & \cnnooave \\
        \ours\ (Ours) 
         & \cnpncr  & \cnpnamzp & \cnpnamaz & \cnpnyelpp & \cnpnyelpf & \cnpnteh & \cnpniave & \cnpnmr & \cnpnsstt & \cnpnsstf & \cnpnoave \\
        \cmidrule(lr){1-12}
        ICL-gold (Oracle) & \cngcr  & \cngamzp & \cngamaz & \cngyelpp & \cngyelpf & \cngteh & \cngiave & \cngmr & \cngsstt & \cngsstf & \cngoave \\
        ICL-random (Oracle)   & \cnrcr  & \cnramzp & \cnramaz & \cnryelpp & \cnryelpf & \cnrteh & \cnriave & \cnrmr & \cnrsstt & \cnrsstf & \cnroave \\
    \midrule
        \multicolumn{11}{l}{\textbf{\em{Direct GPT-NeoX}}} \\
        No-demos & \dnndcr  & \dnndamzp & \dnndamaz & \dnndyelpp & \dnndyelpf & \dnndteh & \dnndiave & \dnndmr & \dnndsstt & \dnndsstf & \dnndoave \\
        Random inputs         & \dnrscr  & \dnrsamzp & \dnrsamaz & \dnrsyelpp & \dnrsyelpf & \dnrsteh & \dnrsiave & \dnrsmr & \dnrssstt & \dnrssstf & \dnrsoave \\
        
        \naive  & \dnnocr  & \dnnoamzp & \dnnoamaz & \dnnoyelpp & \dnnoyelpf & \dnnoteh & \dnnoiave & \dnnomr & \dnnosstt & \dnnosstf & \dnnooave \\
        \ours\ (Ours)
         & \dnpncr  & \dnpnamzp & \dnpnamaz & \dnpnyelpp & \dnpnyelpf & \dnpnteh & \dnpniave & \dnpnmr & \dnpnsstt & \dnpnsstf & \dnpnoave \\
        \cmidrule(lr){1-12}
        ICL-gold (Oracle)              & \dngcr  & \dngamzp & \dngamaz & \dngyelpp & \dngyelpf & \dngteh & \dngiave & \dngmr & \dngsstt & \dngsstf & \dngoave \\
        ICL-random (Oracle)   & \dnrcr  & \dnramzp & \dnramaz & \dnryelpp & \dnryelpf & \dnrteh & \dnriave & \dnrmr & \dnrsstt & \dnrsstf & \dnroave \\
    \bottomrule
    \end{tabular}\vspace{-.1em}
    \caption{
    Results with GPT-J and GPT-NeoX. {\em Oracle} indicates the method has access to the training data, thus is not comparable with the rest of the models.
    Covered/not covered by $\mathcal{C}$ indicates whether or not the domain of the dataset is covered by our text corpus.
    \ours\ is significantly better than previous zero-shot (No-demos) on all datasets, and is on par with ICL-gold on datasets covered by $\mathcal{C}$.
 }\label{tab:main-results}
\end{table*}


\subsection{Experimental Details}\label{subsec:exp-details}

\paragraph{Language models.}
We experiment with three casual language models: GPT-J~\citep{wang2021gpt}, GPT-NeoX~\citep{black2022gpt} and GPT-3~\citep{brown2020language} of sizes 6B, 20B, and 175B, respectively.
We use two inference methods: direct (a regular inference used in \citet{brown2020language}) and channel~\citep{min2021noisy}.

\vspace{-.25em}
\paragraph{Similarity function.}
We define a similarity function $s$ to be a cosine similarity between two sentence embeddings obtained through SimCSE~\citep{tianyu2021simcse}.\footnote{In our initial experiments, we explored multiple embedding methods and found SimCSE works the best.}

\vspace{-.25em}
\paragraph{Implementation details.}
For GPT-J and GPT-NeoX, we use 5 random seeds and report an average and standard deviation.
For GPT-3, we use 2 random seeds and only evaluate on five datasets (CR, Amz, Yelp, Tweet, and SST2) due to limited access.
If the dataset includes more than 2,000 test examples, we subsample 2,000 examples uniformly at random without replacement due to limited computing resources, following prior work~\citep{zhao2021calibrate}.
We use $k=16$ for all experiments.
We use minimal templates from \citet{zhao2021calibrate} without template engineering, e.g., prepending \texttt{Review:} and \texttt{Sentiment:} to the input and the label, respectively, on a review sentiment classification dataset. 
More details are provided in Appendix~\ref{app:impl-details}.

\section{Experimental Results}\label{sec:exp-results}\subsection{Main results}\label{subsec:main-results}

Results using GPT-J and GPT-NeoX are reported in Table~\ref{tab:main-results}.
No-demos outperforms the majority baseline but lags behind ICL-gold or ICL-random that access the training data, confirming the previous work.
Constructing the \pdemos\ using the text corpus significantly helps, e.g., even the ``Random inputs'' baseline is consistently better than No-demos, likely because it informs the label space and the format to the LM.
\naive\ is better than No-demos in many cases but is still worse than ICL-gold.
Finally, \ours, our proposed method, significantly outperforms all baselines.
\ours\ improves zero-shot performance by 5--30\% absolute over the existing zero-shot method (No-demos), consistently over all datasets and all LMs.

\vspace{-.2em}
\paragraph{Comparison to few-shot ICL.}
Compared to oracle baselines that access the training data (ICL-gold and ICL-random), \ours\ performs on par on datasets covered by $\mathcal{C}$, despite being zero-shot. This is fairly consistent over all datasets and LMs. 

On datasets that are not covered by $\mathcal{C}$, \ours\ still lags behind ICL-gold and ICL-random.
This indicates the importance of the coverage of $\mathcal{C}$ in building high-quality \pdemos.
In Section~\ref{subsec:ablations}, we show improving the coverage of $\mathcal{C}$ improves performance on these datasets.

\begin{table*}[ht!]
    \centering \scriptsize
    \setlength{\tabcolsep}{3pt}
    \begin{tabular}{l rrrrar
    }
    \toprule
        \multirow{2}{*}{Method}  & \multicolumn{5}{c}{Covered by $\mathcal{C}$} & Not covered by $\mathcal{C}$ \\
        \cmidrule(lr){2-6} \cmidrule(l){7-7} 
        & CR & Amz & Yelp & Tweet & Avg. & SST-2 \\
    \midrule
        Majority & \rgcr & \rgamzp & \rgyelpp & \rgteh & \rgfave & \rgsstt   \\
    \midrule
        \multicolumn{7}{l}{\textbf{\em{Channel GPT-3}}} \\
        No-demos  & \ctndcr  & \ctndamzp & \ctndyelpp & \ctndteh& \ctndave & \ctndsstt  \\
        \ours\ (Ours)  & \ctpncr  & \ctpnamzp & \ctpnyelpp & \ctpnteh & \ctpnave & \ctpnsstt  \\
        ICL-gold (Oracle)   & \ctgcr  & \ctgamzp & \ctgyelpp & \ctgteh & \ctgave & \ctgsstt  \\
        ICL-random (Oracle)   & \ctrcr  & \ctramzp & \ctryelpp & \ctrteh & \ctrave & \ctrsstt  \\
    \midrule
        \multicolumn{7}{l}{\textbf{\em{Direct GPT-3}}} \\
        No-demos  & \dtndcr & \dtndamzp & \dtndyelpp & \dtndteh & \dtndave & \dtndsstt  \\
        \ours\ (Ours)   & \dtpncr & \dtpnamzp & \dtpnyelpp & \dtpnteh & \dtpnave & \dtpnsstt  \\
        ICL-gold (Oracle)  & \dtgcr  & \dtgamzp & \dtgyelpp & \dtgteh  & \dtgave & \dtgsstt \\
        ICL-random (Oracle)   & \dtrcr  & \dtramzp & \dtryelpp & \dtrteh & \dtrave & \dtrsstt  \\
    \bottomrule
    \end{tabular}\vspace{-.1em}
    \caption{
    Results on GPT-3 on a subset of evaluation datasets.
    %
    %
    {\em Oracle} indicates the method has access to the training data, thus is not comparable with the rest of the model.
    Covered/not covered by $\mathcal{C}$ indicates whether or not the domain of the dataset is covered by our text corpus.
    %
    %
    \ours\ is consistently better than the previous zero-shot (No-demos) on all datasets, even with a template. 
    }\label{tab:main-results-gpt3}
\end{table*}

\vspace{-.2em}
\paragraph{Results with GPT-3.}
Results on a subset of datasets are reported in Table~\ref{tab:main-results-gpt3}.
We find that the findings with GPT-J and GPT-NeoX mostly hold with GPT-3:
\ours\ outperforms the previous zero-shot method (No-demos), and works on par with ICL-gold or ICL-random on datasets covered by $\mathcal{C}$.

\subsection{Ablations}\label{subsec:ablations}

We perform detailed ablation studies that break down the importance of each component of \ours. We evaluate on a subset of 6 datasets (CR, Amz5, Yelp5, Tweet, MR, and SST2) with channel GPT-J unless specified otherwise. 

\begin{figure}
\centering \footnotesize
\resizebox{\columnwidth}{!}{\includegraphics[trim={0.7cm 0.65cm 0.8cm 0.5cm},clip]{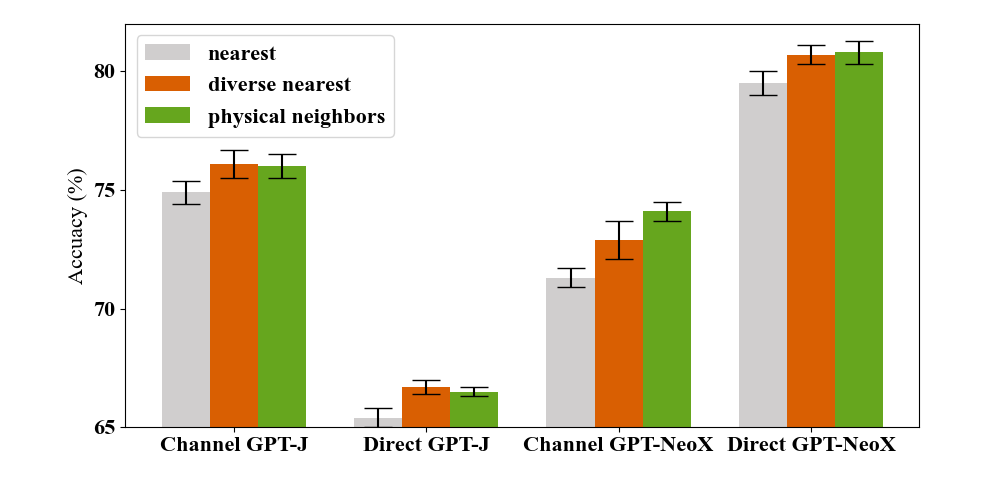}}
\caption{
    \textbf{Effect of the retrieval method.}
    Performance of \ours using different retrieval methods.
    \pn\ is the best retrieval method across different LMs, indicating that it presumably reduces the copying effect the most.
}\label{fig:ablation-retrieval-method}
\end{figure}

\vspace{-.2em}
\paragraph{Effect of the retrieval methods.}
We experiment and compare three different retrieval methods.
(1) \textbf{\topk}, a naive retrieval method that directly selects nearest neighbors $\mathcal{N}_k(x)$ as $x_1, x_2...x_k$. 
(2) \textbf{\topsample}, which first retrieves $K$ nearest neighbors with $x$, $\mathcal{N}_K(x)$, where $K \gg k$, then uniformly samples a random set of $k$ sentences from $\mathcal{N}_K(x)$ as $x_1, x_2...x_k$.\footnote{We use $K=4,096$. 
}
(3) \textbf{\pn}, our main retrieval method introduced in Section~\ref{subsec:method-retrieval}. We do not claim these three methods as the exhaustive set of potential retrieval methods. 

\begin{figure}
\centering \footnotesize
\resizebox{0.8\columnwidth}{!}{\includegraphics[trim={1.1cm 0.7cm 1.2cm 0.5cm},clip]{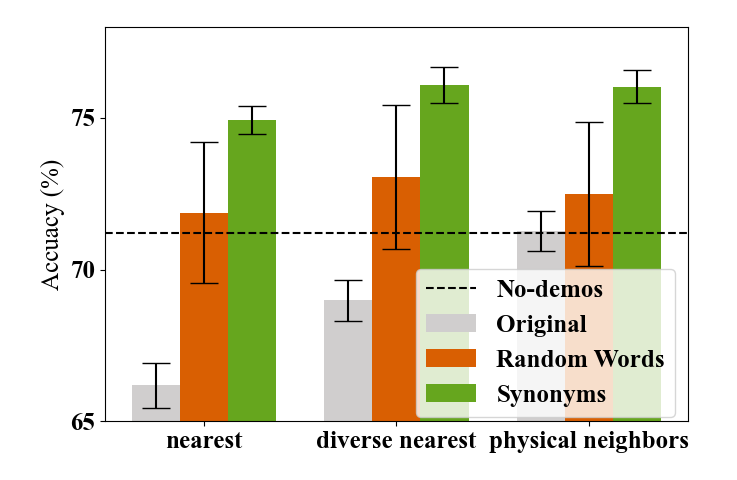}}
\caption{
    \textbf{Effect of synonyms labeling.}
    {\em Original}, {\em Random words}, and {\em Synonyms} indicate the original test labels, random words, and synonyms of the test labels are used in the demonstrations.
    Synonym labeling is critical over all retrieval methods. 
}\label{fig:ablation-synonym}
\end{figure}

Figure~\ref{fig:ablation-retrieval-method} indicates that both `\pn' and `\topsample'  perform well and `\topk' performs the worst consistently over all LMs. 
This indicates that while informing the input space of the test input, encouraging more diversity in the \pdemos\ is important, presumably because they are more effective in reducing the copying effect. 

\begin{figure*}
\centering \footnotesize
\resizebox{1.7\columnwidth}{!}{\includegraphics[trim={4.4cm 0.8cm 3.2cm 0.7cm},clip]{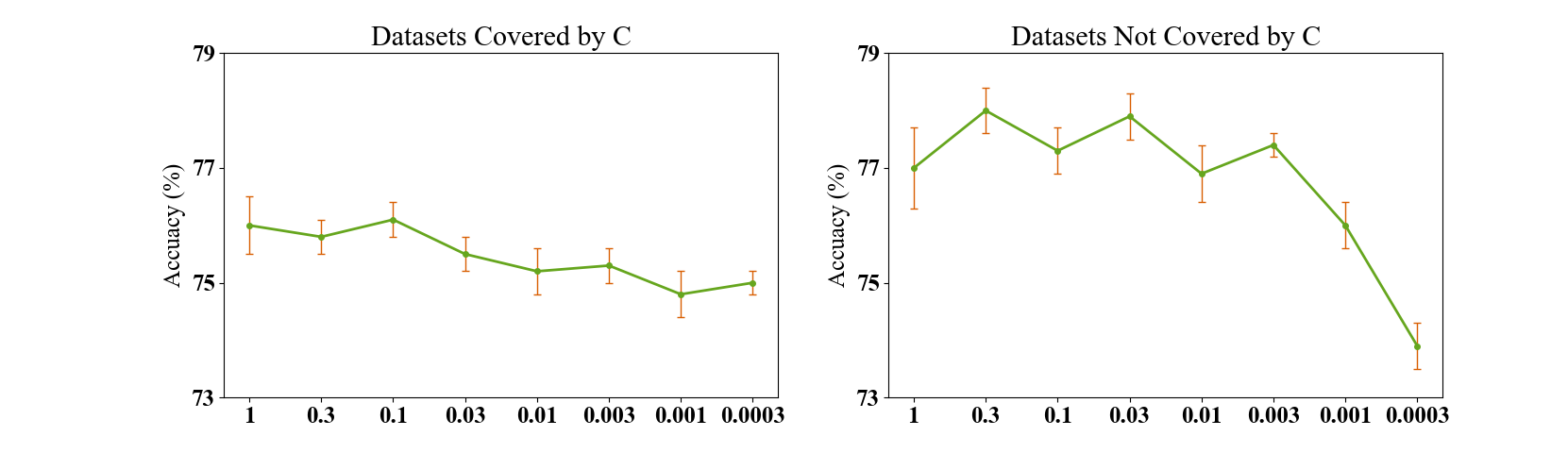}}
\caption{
    \textbf{Effect of the size of the corpus.}
    The $x$-axis indicates the size of the corpus, varying from 160M paragraphs (1) to 48K paragraphs (0.0003).
    Performance goes down as the corpus size decreases.
}\label{fig:ablation-size}
\end{figure*}

\begin{figure}
\centering \footnotesize
\resizebox{0.9\columnwidth}{!}{\includegraphics[trim={0.6cm 0.6cm 0.3cm 0.6cm},clip]{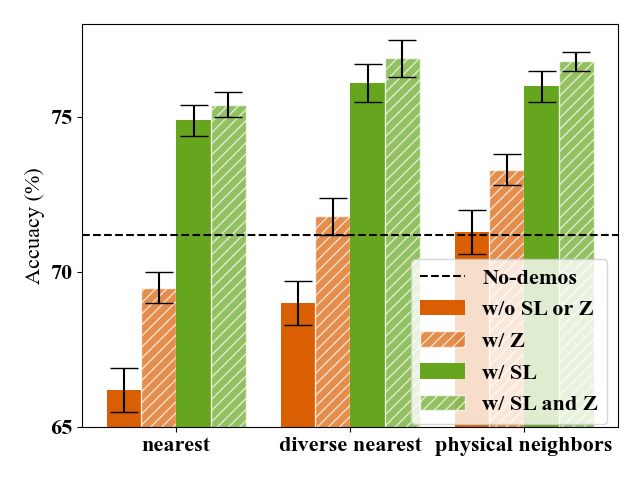}}
\caption{
    \textbf{Quantifying the Copying Effect.}
    \em{SL} and \em{Z} stand for synonym labeling and zeroing out the attention heads, respectively.
    Techniques for reducing the copying effect (physical neighbor and synonym labeling) are less affected by zeroing out the attention heads.
}\label{fig:ablation-avoidng-copying}
\end{figure}

\vspace{-.2em}
\paragraph{Effect of synonym labeling.}
We aim to answer two questions: (a) How is the effect of synonym labeling when different retrieval methods are used? (b) How important is it to keep the semantics of the label words, e.g., what if we use random words instead of synonyms?
To answer these questions, we compare three different methods of assigning labels: (1) using the original test labels, (2) using random words,\footnote{We construct a 1-1 mapping between the original test labels and random English unigrams, and assign the labels. Thus, the number of unique words used in the \pdemos\ is the same as the number of unique labels.}
and (3) using the synonyms of the test labels, over the three different retrieval methods.

Results are shown in Figure~\ref{fig:ablation-synonym}.
Using random words is consistently better than using the original labels, indicating that not using words from original test labels is important.
Nonetheless, using synonyms is consistently better than using random words, indicating that informing the semantic space of the labels is still important.
While these trends are consistent across different retrieval methods, the gap between using the original labels and using the synonyms is smaller when the retrieval method encourages diversity, e.g., the smallest with the \pn\ method and the largest with the \topk\ method.
This is likely because the \pn\ method is already partially reducing the copying effect.



\vspace{-.2em}
\paragraph{Quantifying the Copying Effect.}
To better quantify how much the gains are from avoiding the copying effect, we follow \citet{anonymous2023overthinking} in (1) identifying some attention heads in the Transformer layers that are most responsible for copying, and (2) zero-ing their weights out.
If this leads to performance improvements, it is a strong indicator that the method has been suffering from the copying effect.
%
%
We apply this method to three different retrieval methods: \topk, \topsample\, and \pn\ introduced in Section~\ref{subsec:method-retrieval}.

Figure~\ref{fig:ablation-avoidng-copying} reports results.
First, all methods have performance improvements by zero-ing out the attention heads, indicating that all of them suffer from the copying effect to a certain degree. 
We then find that (1) \topk\ is affected the most and \pn\ is affected the least, and (2) methods with synonym labeling are affected much less than their counterpart without synonym labeling.
These are aligned with our earlier intuition that using \pn\ instead of \topk, and using synonym labeling help reducing the copying effect.


\vspace{-.2em}
\paragraph{Effect of the size of the corpus.}
We quantify the impact of the size of the corpus.
This is important to judge whether \ours\ can potentially achieve better results by scaling the corpus.
We evaluate \ours\ with a corpus with varying sizes, from 100\% to 0.03\% of the corpus.

Figure~\ref{fig:ablation-size} demonstrates that performance goes down as the size of the corpus gets smaller.
This is likely because there are less sentences that are sufficiently close to the test input when the corpus is smaller, thus the {\em relevance} of the nearest neighbors and the test input drops.
This trend is clearer on the \iddatasets\ than on the \ooddatasets.

\vspace{-.2em}
\paragraph{Effect of the format of demonstrations.}
How many input-label pairs does \ours\ need to benefit from \pdemos?
Are gains from \pdemos\ mainly from the fact that the LM conditions on relevant text, or does the LM benefit from a specific format of the \pdemos: a concatenation of input-label pairs?
To answer these questions, we experiment with (1) \ours\ with varying range of $k$ from $1$ to $64$, and (2) a variant of \ours\ where the LM conditions on a concatenation of retrieved inputs, without randomly paired labels (called ``Inputs-only'').

\begin{figure}
\centering \footnotesize
\resizebox{\columnwidth}{!}{\includegraphics[trim={1.5cm 0.5cm 3cm 1cm},clip]{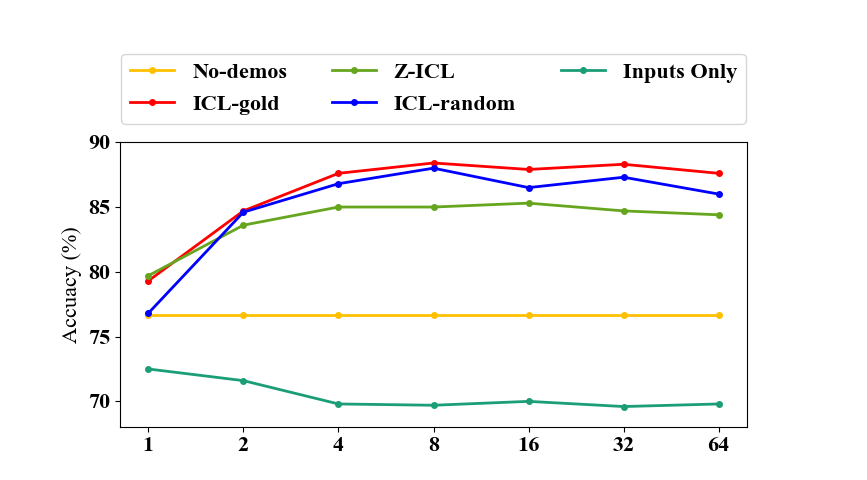}}
\caption{
    \textbf{Effect of the format of demonstrations} with varying numbers of demonstrations ($k$).
    \ours\ consistently performs on par with the oracle baseline, and ``Inputs-only'' performs significantly worse.
}\label{fig:ablation-form}
\end{figure}

Results are shown in Figure~\ref{fig:ablation-form}.
First, \ours\ is significantly better than zero-shot baselines and stays on par with the oracle baselines consistently across different values of $k$.
Moreover, using no labels (``Inputs-only'') performs significant worse than its counterparts.
This suggests that \ours\ takes advantages of the form of input-label pairs, and is beyond simply conditioning on relevant context.

\vspace{-.2em}
\paragraph{Effect of the coverage of the corpus.}
We quantify the impact of the coverage of the corpus, and whether adding more domains in the corpus improves performance.
We do so by adding the unlabeled portion of IMDB review~\citep{mass2011imdb} to the corpus $\mathcal{C}$.
The size of $\mathcal{C}$ increases only by 2\%, but covers the domain of three datasets that were previously not covered (SST2, SST5 and MR).

\begin{figure}
\centering \footnotesize
\resizebox{0.8\columnwidth}{!}{\includegraphics[trim={0.5cm 0.5cm 0.3cm 0.5cm},clip]{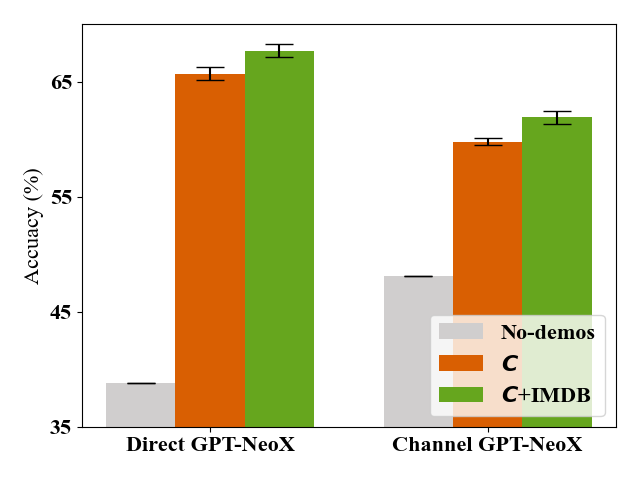}}
\caption{
    \textbf{Effect of the coverage of the corpus.}
    Performance of \ours\ before and after IMDB is added to the corpus.
    Expanding the coverage of the corpus consistently improves the performance despite only 2\% of the increase in the size of the corpus.
}\label{fig:ablation-coverage}
\end{figure}

Figure~\ref{fig:ablation-coverage} shows the performance on three datasets before and after adding the IMDB corpus.
Performance improves consistently over all LMs, even though it only adds up the size by 2\%.
This suggests that the coverage of the text corpus is important, and it is feasible to further improve the overall performance simply by expanding the corpus.

\section{Conclusion}\label{sec:concl}

We introduced \ours, a zero-shot in-context learning method that constructs \pdemos\ from a raw text corpus.
Our method (1) retrieves relevant text from the corpus using the nearest neighbor search, effectively informing the correct space of the inputs to the LM, and (2) adjust the \pdemos\ with \pn\ and synonym labeling to avoid the copying effect.
Evaluation on nine classification datasets shows
\ours\ significantly outperforms the previous zero-shot baseline, and performs on par with the $k$-shot \demos.
Overall, \ours\ demonstrates that significantly higher LM zero-shot performance is possible, and opens up a new research direction on the construction of better \pdemos\ that expose the full capacity of a LM.

\section*{Limitation}\label{sec:lim}

\paragraph{Extension to multi-sentence tasks.}
Our experiments are limited to single-sentence tasks, as we only retrieve single-sentence nearest neighbors to a test input.
Multi-sentence tasks such as natural language inference would require constructing \pdemos\ that consists of multiple sentences, which we leave for future work. 

\vspace{-.2em}
\paragraph{Beyond classification.}
Our experiments are limited to classification.
Extensions to multi-choice tasks or generation tasks requires going beyond a fixed set of options shared between inputs in the \demos\ and the test input. We leave extensions to non-classification tasks for future work.

%
\vspace{-.2em}
\paragraph{Better construction of \pdemos.}
We think future work can explore better constructing the \pdemos.
For instance, this paper uses manually chosen synonym labels (see Appendix~\ref{app:impl-details} for more detail).
We hypothesize that better \pdemos\ can improve performance, which we leave for future work. 

\section*{Acknowledgements}
We thank UW NLP members and anonymous reviewers for their comments in the paper.
This research was supported by NSF IIS-2044660, an Allen Distinguished Award and gifts from AI2. SM is supported by a J.P. Morgan fellowship.

\bibliography{abbr,acl}
\bibliographystyle{acl_natbib}

\clearpage
\appendix
\section{Data Statistics}\label{app:statistics}

\begin{table*}[!b]
    \centering \myfontsize 
    \begin{tabular}{l @{\hspace{2em}} l r}
        \toprule
            Domain & Description & \#sentences\\
        \midrule
            1B      & NewsWire sentences  & 1.0M\\ 
            CS      & full-text CS papers from S2ORC &  1.0M \\
            LEGAL   & U.S. court opinions, 1658 to 2018 &  3.0M   \\
            MED   & full-text medical papers from S2ORC & 1.0M \\
            WEBTEXT   & Web documents &  2.1M  \\
            REALNEWS   & articles from REALNEWS & 1.8M    \\
            REDDIT   & Reddit comments from pushshift.io & 2.6M \\
            REVIEWS   & Amazon product reviews & 3.1M \\
            ACL PAPERS   & NLP papers from ACL &   46K   \\
            BREAKING NEWS   & latest articles from 400 English news sites & 0.5M    \\
            CONTRACTS   & commercial legal contracts & 47K    \\
            CORD-19   & excerpts from COVID-19 research papers &  0.9M   \\
            GITHUB   & public Github repository contents &   0.6M   \\
            GUTENBERG   & copyright-expired books & 0.9M     \\
            TWEETS   & English tweets from 2013-2018 &  0.8M    \\
            YELP REVIEWS   & Yelp restaurant reviews &  7.5M    \\
        \bottomrule
    \end{tabular}\vspace{-.1em}
    \caption{
    List of domains
    from \citet{suchin2021demix}.
    }\label{tab:corpuses}
\end{table*}

\begin{table*}[!b]
    \centering  \myfontsize 
    \begin{tabular}{l @{\hspace{1em}} c @{\hspace{1.5em}} l @{\hspace{1em}} l}
        \toprule
            Dataset & \# examples & labels & synonyms\\
        \midrule
             \multicolumn{4}{l}{~~~~\textbf{\em \small Datasets covered by $\mathcal{C}$}} \\
            CR      & 2,000 &  "terrible", "great"  & "bad", "good" \\
            Amz      & 1,000 & "negative", "positive"  & "bad", "good" \\
            Amz5   & 100,050 $\rightarrow$ 2,000 & "terrible", "bad", "okay", "good", "great" & "horrible", "negative", "neutral", "positive", "excellent" \\
            Yelp   & 7,600 $\rightarrow$  2,000 & "negative", "positive"  &  "bad", "good" \\
            Yelp5   & 50,000 $\rightarrow$ 2,000 & "terrible", "bad", "okay", "good", "great" &  "horrible", "negative", "neutral", "positive", "excellent" \\
            Tweet   & 2,000 & "negative", "neutral", "positive"  &  "bad", "normal", "good"\\
            \cmidrule(lr){1-4}
            \multicolumn{4}{l}{~~~~\textbf{\em \small Datasets not covered by $\mathcal{C}$}} \\
            MR   &  2,000  &"terrible", "great"  &  "bad", "good" \\
            SST2   &872& "terrible", "great" &  "bad", "good" \\
            SST5   & 2,210 $\rightarrow$ 2,000  &"terrible", "bad", "okay", "good", "great" & "horrible", "negative", "neutral", "positive", "excellent" \\
        \bottomrule
    \end{tabular}\vspace{-.1em}
    \caption{
        Statistics of evaluation datasets as well as their labels and synonyms.
    }\label{tab:eval-datasets}
\end{table*}

\paragraph{Corpus.} We take the same English corpus from \cite{suchin2021demix} covering 16 diverse domains: 1B, CS, LEGAL, MED, WEBTEXT, REALNEWS, REDDIT, REVIEWS, ACL PAPERS, BREAKING NEWS, CONTRACTS, CORD-19, GITHUB, GUTENBERG, TWEETS, and YELP REVIEWS. See the descriptions and statics in Table~\ref{tab:corpuses}. For each domain, we 1) subsample 10M paragraphs if the data is larger, 2) split each paragraph into sentences, and 3) remove duplicate sentences while keeping the ordering of the sentences as in the original paragraphs. 

\paragraph{Evaluation datasets.}
Statistics and descriptions of our evaluation datasets are reported in Table~\ref{tab:eval-datasets}. For each dataset, we subsample 2000 test examples uniformly at random if the test data is larger, due to limited computational resources.

\section{Implementation Details}\label{app:impl-details}

All implementations are done in PyTorch~\citep{paszke2019pytorch}.
We use int8 quantization \citep{zeng2022glm} to run GPT-NeoX on 40GB A100 machines.

\paragraph{Format of the demonstrations.}
We use $k=16$ demonstration examples for all the baselines and methods, unless specified otherwise.
We truncate each demonstration example to have up to 256 tokens and the concatenation of them to have up to 1,024 tokens.

\paragraph{Nearest neighbor search.} We use SimCSE~\citep{tianyu2021simcse} to embed the corpus and the test inputs. 
We use FAISS~\citep{johnson2019billion} to build an index for the corpus offline and perform nearest neighbor search at inference.

\paragraph{Synonym labeling.}
We manually choose a synonym of each label to perform synonym labeling. 
A full list of synonyms is reported in Table~\ref{tab:eval-datasets}.

\paragraph{Computational Budget.} Our main experiment on the 4 public LMs in Table~\ref{tab:main-results} takes around 4,000 computing hours with a 40GB A100 machine.
Our experiment using GPT-3's API
costs around 4,500 US Dollars.

\paragraph{}

\end{document}